\documentclass{article}

\usepackage{microtype}
\usepackage{graphicx}
\usepackage{subcaption}
\usepackage{booktabs} % for professional tables
\usepackage{hyperref}
\usepackage{url}
\usepackage{amsmath}
\usepackage{amssymb}
\usepackage{soul}

\newcommand{\method}{FlowGMM\xspace}
\newcommand{\methodcons}{FlowGMM-cons\xspace}
\newcommand{\mX}{\mathcal{X}}
\newcommand{\mC}{\mathcal{C}}
\newcommand{\mZ}{\mathcal{Z}}
\newcommand{\mN}{\mathcal{N}}

\usepackage{graphicx}
\usepackage{graphbox}
\usepackage{booktabs}
\usepackage{subcaption}
\usepackage{xspace}

\usepackage{tikz}
\usepackage[symbol]{footmisc}

\usepackage{hyperref}

\usepackage[accepted]{icml2019}

\usepackage{xcolor}
\definecolor{dark-red}{rgb}{0.4,0.15,0.15}
\definecolor{dark-blue}{rgb}{0.15,0.15,0.4}
\definecolor{medium-blue}{rgb}{0,0,0.5}
\hypersetup{
   colorlinks, linkcolor={dark-blue},
   citecolor={dark-blue}, urlcolor={medium-blue}
}

\icmltitlerunning{Semi-Supervised Learning with Normalizing Flows}

\begin{document}

\twocolumn[
\icmltitle{Semi-Supervised Learning with Normalizing Flows}

\icmlsetsymbol{equal}{*}

\begin{icmlauthorlist}
\icmlauthor{Pavel Izmailov}{equal,nyu}
\icmlauthor{Polina Kirichenko}{equal,nyu}
\icmlauthor{Marc Finzi}{equal,nyu}
\icmlauthor{Andrew Gordon Wilson}{nyu}
\end{icmlauthorlist}

\icmlaffiliation{nyu}{New York University}

\icmlcorrespondingauthor{Pavel Izmailov}{pi390@nyu.edu}
\icmlcorrespondingauthor{Andrew Gordon Wilson}{andrewgw@cims.nyu.edu}

\icmlkeywords{Semi-supervised learning, generative models, Bayes classifier}

\vskip 0.3in
]

\printAffiliationsAndNotice{\icmlEqualContribution} 

\begin{abstract}
Normalizing flows transform a latent distribution through an invertible neural network for a flexible and pleasingly simple approach to generative modelling, while preserving an exact likelihood. We propose \method, an end-to-end approach to generative semi-supervised learning with normalizing flows, using a latent Gaussian mixture model. \method is distinct in its simplicity, unified treatment of labelled and unlabelled data with an exact likelihood, interpretability, and broad applicability beyond image data. We show promising results on a wide range of applications, including AG-News and Yahoo Answers text data, tabular data, and semi-supervised image classification. We also show that \method can discover interpretable structure, provide real-time optimization-free feature visualizations, and specify well calibrated predictive distributions.

\end{abstract}

\section{Introduction}

The discriminative approach to classification models
the probability of a class label given an input $p(y|x)$ directly. The generative approach,
by contrast, models the class conditional density for the data $p(x|y)$, and then uses 
Bayes rule to find $p(y|x)$. In principle, generative modelling has long been more 
alluring, for the effort is focused on creating an interpretable object of 
interest, and ``what I cannot create, I do not understand''. In practice, 
discriminative approaches typically outperform generative methods, and thus are far more widely used.

The challenge in generative modelling is that standard approaches to density estimation
are often poor descriptions of high-dimensional natural signals. For example, a Gaussian mixture
directly over images, while highly flexible for estimating densities, would specify similarities
between images as related to Euclidean distances of pixel intensities, which would be a poor
inductive bias for handling translations or representing other salient statistical properties. Recently, generative adversarial networks \citep{goodfellow2014generative}, variational autoencoders \citep{kingma2013auto}, and normalizing flows \citep{dinh2014nice}, have 
led to great advances in unsupervised generative modelling, by leveraging the inductive biases of deep convolutional neural networks.

Normalizing flows are a pleasingly simple approach to generative modelling, which work by transforming a distribution through an invertible neural network. Since the transformation is invertible, it is possible to exactly express the likelihood over the observed data, to train the neural network mapping. The network provides both useful inductive biases, and a flexible approach to density estimation. Normalizing flows
also admit controllable latent representations and can be sampled efficiently, unlike auto-regressive models \citep{papamakarios2017masked, oord2016pixel}. 
Moreover, recent work \citep{dinh2016density,kingma2018glow,behrmann2018invertible, chen2019residual, song2019mintnet} demonstrated that normalizing flows can produce high-fidelity samples for natural image datasets.

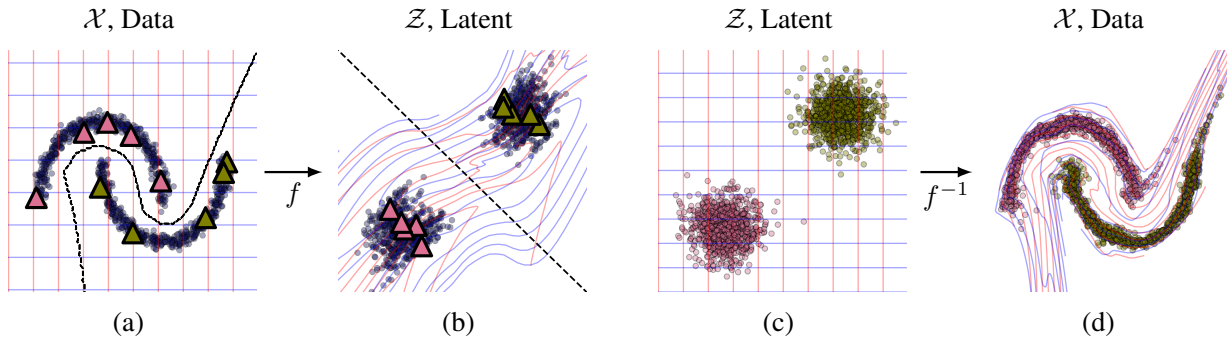
\begin{figure*}
\resizebox{1.0\textwidth}{!}{

\hspace{-0.5cm}
\begin{tikzpicture}

\node (a) {\includegraphics[width=4.cm]{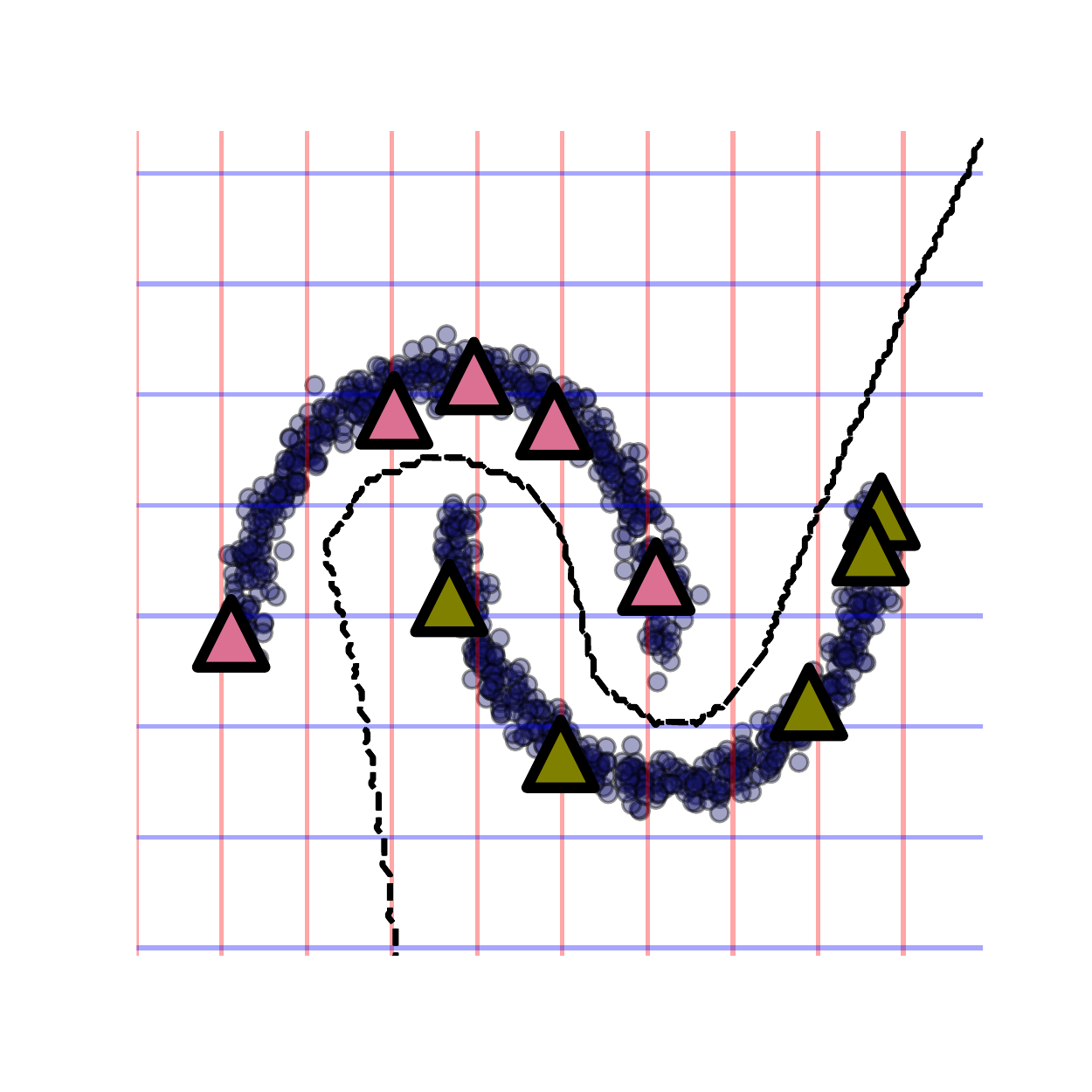}};
\node (b)[right=2.cm] {\includegraphics[width=4.cm]{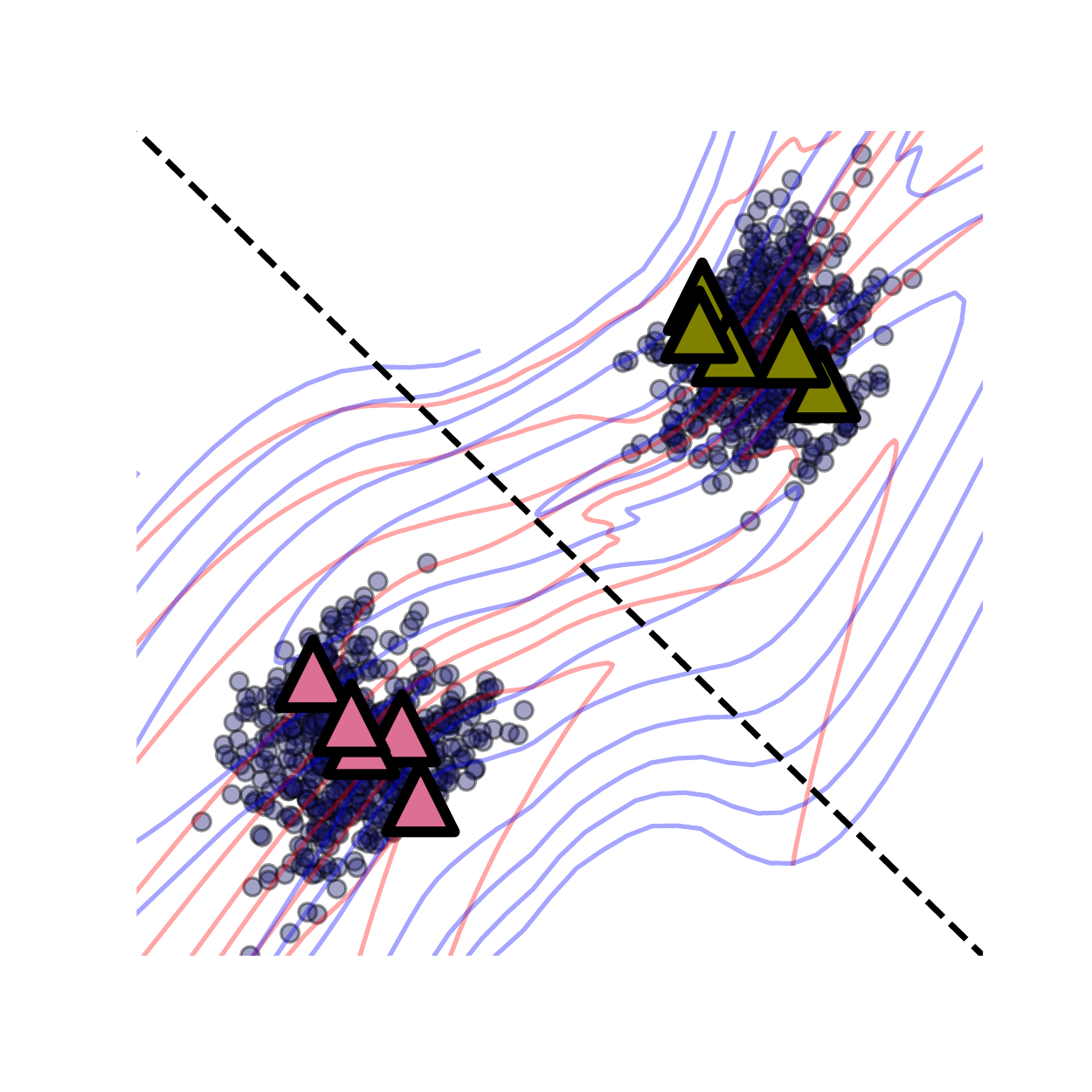}};
\node (c)[right=6.cm] {\includegraphics[width=4.cm]{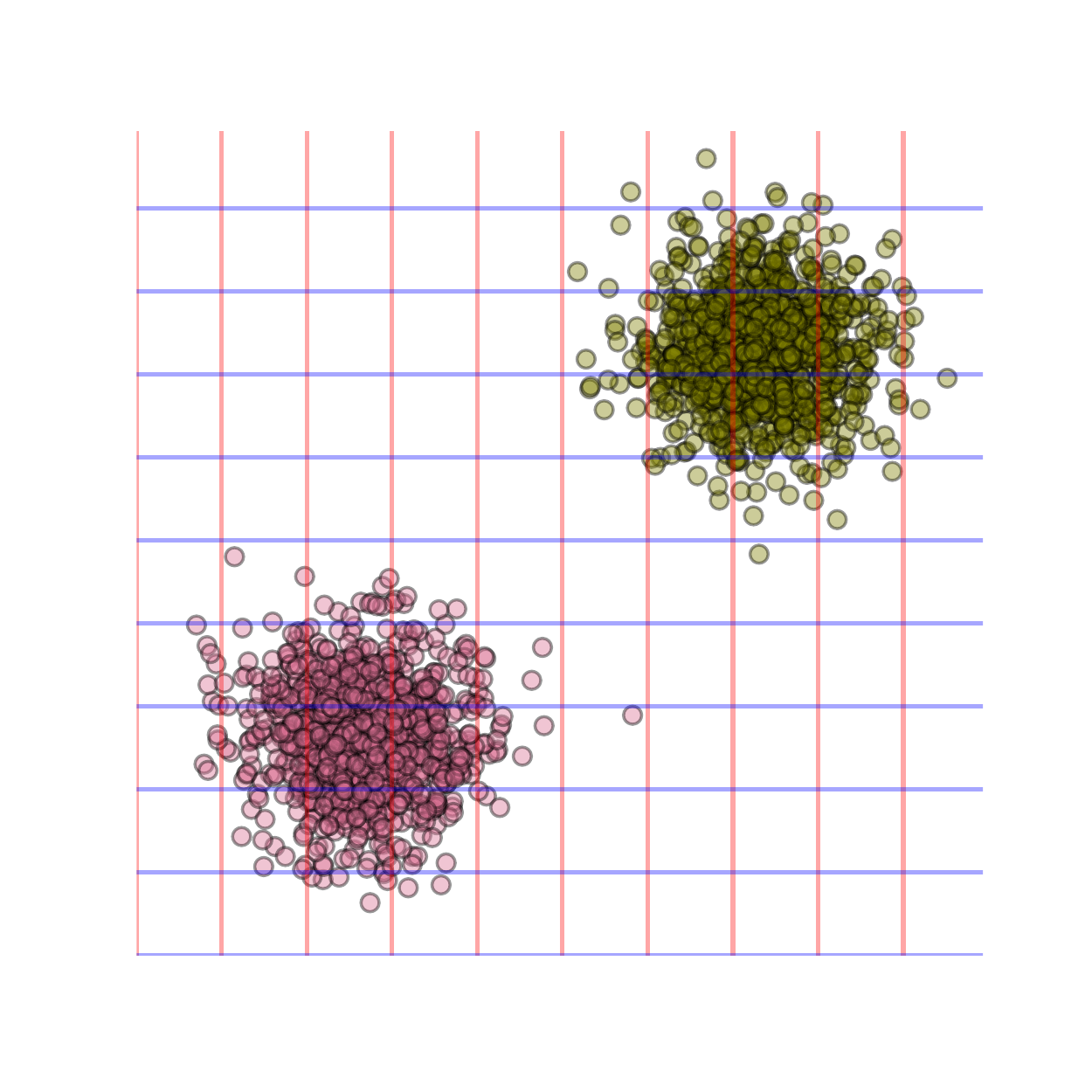}};
\node (d)[right=10.cm] {\includegraphics[width=4.cm]{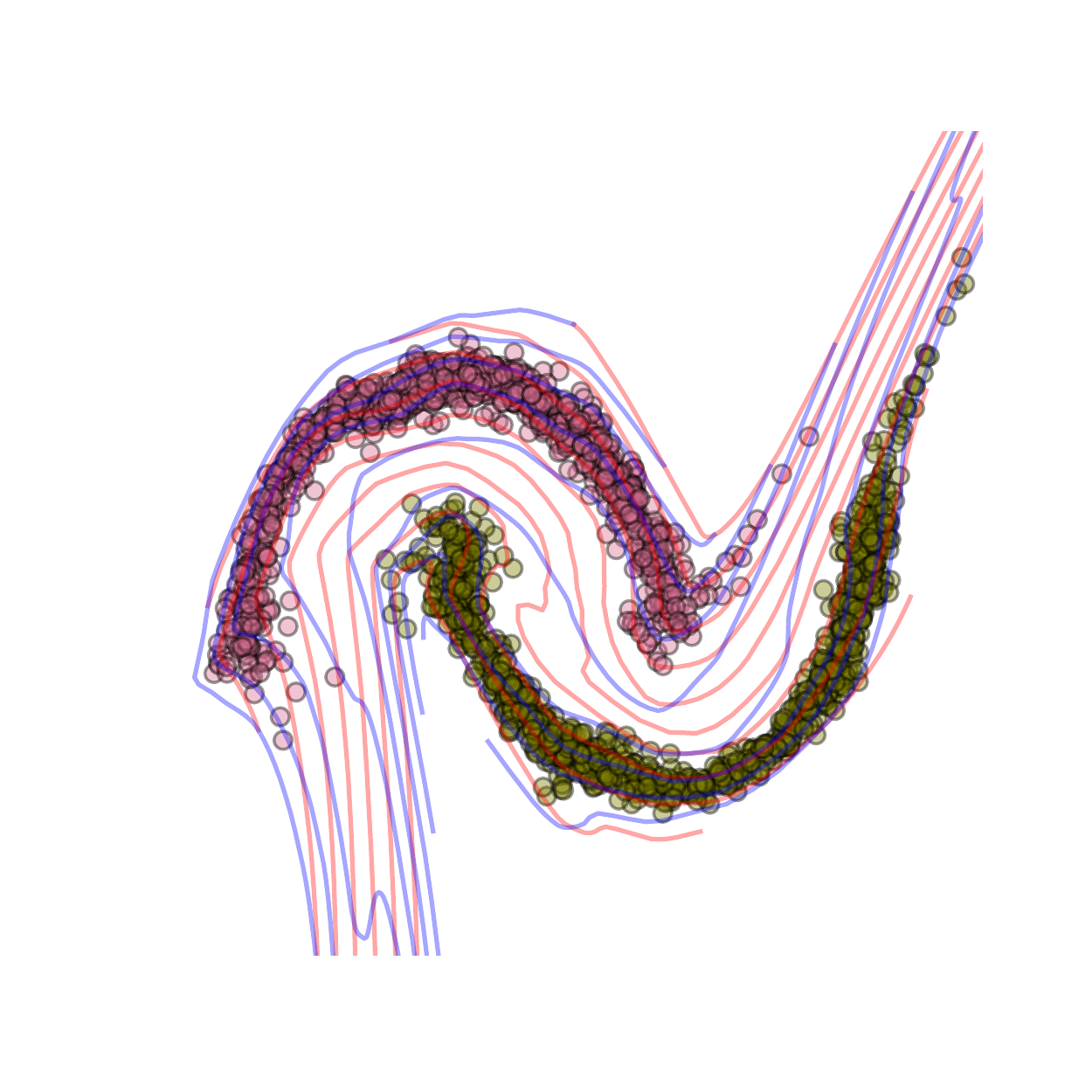}};

\draw[-latex,thick] ([xshift=-3mm] b.west) -- ([xshift=3mm] a.east) 
node[midway,below]{$f$};
\draw[-latex,thick] ([xshift=-1mm] d.west) -- ([xshift=3mm] c.east) 
node[midway,below]{$f^{-1}$};

\draw (a.north) node[above=-5mm]{$\mX$, Data};
\draw (b.north) node[above=-5mm]{$\mZ$, Latent};
\draw (c.north) node[above=-5mm]{$\mZ$, Latent};
\draw (d.north) node[above=-5mm]{$\mX$, Data};
\draw (a.south) node[below=-5mm]{(a)};
\draw (b.south) node[below=-5mm]{(b)};
\draw (c.south) node[below=-5mm]{(c)};
\draw (d.south) node[below=-5mm]{(d)};

\end{tikzpicture}
}
\caption{
        Illustration of semi-supervised learning with \method on a binary classification problem. 
        Colors represent the two classes or the corresponding Gaussian mixture components.
        Labeled data are shown with triangles, colored by the corresponding class label, and 
        blue dots represent unlabeled data.
        \textbf{(a):} Data distribution and the classifier decision boundary.
        \textbf{(b):} The learned mapping of the data to the latent space.  
        \textbf{(c):} Samples from the Gaussian mixture in the latent space.
        \textbf{(d):} Samples from the model in the data space.
    }
	\label{fig:intro}
\end{figure*}

Advances in unsupervised generative modelling, such as normalizing flows, are particularly compelling for
\emph{semi-supervised} learning, where we wish to share structure over labelled and unlabelled data, to make better predictions of class labels on unseen data. In this paper, we introduce an approach to semi-supervised learning with normalizing flows, by modelling the density in the \emph{latent space} as a Gaussian mixture, with each mixture component corresponding to a class represented in the labelled data. This \emph{Flow Gaussian Mixture Model} (\method) is to the best of our knowledge the first approach to semi-supervised learning with normalizing flows that provides an exact joint likelihood over both labelled and unlabelled data, for end-to-end training.\footnote{A short version of this work first appeared at the ICML 2019 Normalizing Flows Workshop \citep{izmailov2019semi}. At the same workshop, \citet{atanov2019semi} proposed a different approach that uses a class-conditional normalizing flow as the latent distribution.}

We illustrate \method with a simple example in Figure \ref{fig:intro}.
We are solving a binary semi-supervised classification problem on the dataset shown 
in panel (a): the labeled data are shown with triangles colored according to their class, 
and unlabeled data are shown with blue circles. 
We introduce a Gaussian mixture with two components corresponding to each of the classes,
shown in panel (c) in the latent space $\mZ$ and an invertible transformation $f$.
The transformation $f$ is then trained to map the data distribution in the data space $\mX$ to the latent Gaussian mixture in the $\mZ$ space, mapping the labeled data to the corresponding mixture component.
We visualize the learned transformation in panel (b), showing the positions of the images
$f(x)$ for all of the training data points.
The inverse $f^{-1}$ of this mapping serves as a class-conditional generative model, that we visualize in panel (d).
To classify a data point $x$ in the input space we compute its image $f(x)$ in the latent space, and pick the class corresponding to the Gaussian that is closest to $f(x)$. 
We visualize the decision boundary of the learned classifier with a dashed line in panel (a).

\method naturally encodes the \textit{clustering principle}: the decision boundary
between classes must lie in the low-density region in the data space. 
Indeed, in the latent space
the decision boundary between two classes coincides with the hyperplane perpendicular to the line
segment connecting means of the corresponding mixture components and passing through the midpoint of this line segment (assuming the components are normal distributions with identity covariance matrices); in panel (b) of Figure \ref{fig:intro} we show the decision boundary in the latent space with a dashed line.
The density of the latent distribution near
the decision boundary is low. 
As the flow is trained to represent data as a transformation
of this latent distribution, the density near the decision boundary should
also be low. In panel (a) of Figure \ref{fig:intro} the decision
boundary indeed lies in the low-density region.

The contributions of this work include:

\begin{itemize}

\item We propose \method, a new probabilistic classification model based on 
normalizing flows that can be naturally applied to semi-supervised learning.
\item We show that \method has good performance on a broad range of semi-supervised tasks, including image, text and tabular data classification.
\item We propose a new type of probabilistic consistency regularization that significantly improves \method on image classification problems.  
\item To demonstrate the interpretability of \method, we visualize the learned latent space representations for the proposed semi-supervised model and show that interpolations between data points from different classes pass through low-density regions. We also show how \method can be used for feature visualization in real-time, without requiring gradients.
\item We show that the predictive uncertainties of \method can be naturally calibrated by scaling the variances of mixture components. 
\item We provide code for \method at: \url{https://github.com/izmailovpavel/flowgmm}

\end{itemize}

\section{Related Work}

 \citet{kingma2014semi} represents one of the earliest works on semi-supervised deep generative modelling, demonstrating how the likelihood model of a variational autoencoder \citep{kingma2013auto} could be used for semi-supervised image classification. \citet{xu2017variational} later extended this framework to semi-supervised text classification. 
 
Many generative models for classification \citep{salimans2016improved,nalisnick2019hybrid,chen2019residual} have relied upon multitask learning, where a shared latent representation is learned for the generative model and the classifier.
With the method of \citet{chen2019residual}, hybrid modeling is observed to \textit{reduce} performance for both tasks in the supervised case.
While GANs have achieved promising performance on semi-supervised tasks, \citet{dai2017} showed that classification performance and generative performance are in direct conflict: a perfect generator provides no benefit to classification performance. 

Some works on normalizing flows, such as RealNVP \citep{dinh2016density},
have used class-conditional sampling, where the transformation is conditioned on the class label. These approaches pass the class label as an input to coupling layers, conditioning the output of the flow on the class.

Deep Invertible Generalized Linear Model \citep[DIGLM, ][]{nalisnick2019hybrid}, most closely related to our work, trains a classifier on the latent representation of a normalizing flow
to perform supervised or semi-supervised image 
classification. 
Our approach is principally different, as we
use a mixture of Gaussians in the latent space $\mZ$
and perform classification based on class-conditional
likelihoods (see \eqref{eq:preds}), rather than training a
separate classifier. 
One of the key advantages of our
approach is the explicit encoding of clustering principle in the method and a more natural probabilistic interpretation.

Indeed, many approaches to semi-supervised learn from the labelled and unlabelled data using different (and possibly misaligned) objectives, often also involving two step procedures where the unsupervised model is used as pre-processing for a supervised approach.
In general, \method is distinct in that the generative model is used directly as a Bayes classifier, and in the limit of a perfect generative model the Bayes classifier achieves a provably optimal misclassification rate \citep[see e.g.][]{mohri2018foundations}. Moreover, approaches to semi-supervised classification, such as consistency regularization \citep{laine2016temporal, miyato2018virtual,tarvainen2017mean, athiwaratkun2018there, verma2019interpolation, berthelot2019mixmatch}, typically focus on image modelling. We instead focus on showcasing the broad applicability of \method on text, tabular, and image data, as well as the ability to conveniently discover interpretable structure.

\section{Background: Normalizing Flows}
\label{sec:background}

The normalizing flow \citep{dinh2016density} is an unsupervised model
for density estimation defined as an invertible mapping 
$f: \mX \rightarrow \mZ$ from the data space $\mX$ to the latent space 
$\mZ$.
We can model the data distribution as a transformation 
$f^{-1}: \mZ \rightarrow \mX$ applied to a random 
variable from the latent distribution $z \sim p_\mZ$, which is often chosen to be Gaussian.
The density of the transformed random variable $x = f^{-1}(z)$ is 
given by the change of variables formula
\begin{equation}\label{eq:nf}
    p_\mX(x) = p_\mZ(f(x)) \cdot \left| \det \left( \frac {\partial f}{\partial x} \right) \right|.
\end{equation}
The mapping $f$ is implemented as a sequence of invertible functions, parametrized by a neural network with architecture
that is  designed to ensure invertibility and efficient computation 
of log-determinants, and a set of parameters $\theta$ that can be optimized. The model can be trained by maximizing the likelihood in Equation~\eqref{eq:nf} of the training
data with respect to the parameters $\theta$.

\section{Flow Gaussian Mixture Model}
\label{sec:sslnf}

We introduce the Flow Gaussian Mixture Model (FlowGMM), a probabilistic generative model for semi-supervised learning with normalizing flows.
In \method, we introduce a discrete latent variable $y$ for the class label, $y \in \{1\dots \mC\}$. 
Our latent space distribution, conditioned on a label $k$, is Gaussian with mean $\mu_k$ and covariance $\Sigma_k$:
\begin{equation}\label{eq:conditional}
    p_\mZ(z|y=k) = \mN(z | \mu_k, \Sigma_k).
\end{equation}
The marginal distribution of $z$ is then a Gaussian mixture. When the classes are balanced, this distribution is
\begin{equation}\label{eq:mixture}
    p_\mZ(z) = \frac 1 \mC \sum_{k=1}^\mC \mN(z | \mu_k, \Sigma_k).
\end{equation}
Combining equations \eqref{eq:conditional}  and \eqref{eq:nf}, the likelihood for labeled data is 
\begin{equation*}
     p_\mX(x|y=k) = \mN \left(f(x) | \mu_{k}, \Sigma_{k}\right) \cdot \left| \det \left( \frac {\partial f}{\partial x} \right) \right|,
\end{equation*}
and the likelihood for data with unknown label is $p_\mX(x) = \sum_k p_\mX(x|y=k)p(y=k)$.
If we have access to both a labeled dataset $\mathcal{D}_\ell$ and an unlabeled 
dataset $\mathcal{D}_u$, then we can train our model in a semi-supervised way to maximize the joint likelihood of the labeled and unlabeled data
\begin{equation}\label{eq:loss}
    p_\mX(\mathcal{D}_\ell,\mathcal{D}_u | \theta) = \prod_{(x_i,y_i)\in \mathcal{D}_\ell} p_\mX(x_i,y_i)\prod_{x_j\in \mathcal{D}_u} p_\mX(x_j),
\end{equation}
over the parameters $\theta$ of the bijective function $f$, which learns a density model for a Bayes classifier. In particular, given a test point $x$, the model predictive distribution is given by
\begin{align}
    p_\mX(y|x) &=   \frac{p_\mX(x|y)p(y)}{p(x)} \notag \\ &= 
    \frac{\mN \left(f(x) | \mu_{y}, \Sigma_{y}\right)}{\sum_{k=1}^{\mC} \mN(f(x) \vert \mu_k, \Sigma_k)}. 
   \label{eq:preds}
\end{align}
We can then make predictions for a test point $x$ with the Bayes decision rule $${y = \arg \max_{i\in\{1, \ldots, \mC\}}p_\mX(y=i|x)}.$$

As an alternative to direct likelihood maximization, we can adapt the Expectation Maximization algorithm for model training as discussed in Appendix \ref{sec:em}.

\begin{figure*}[t]
	\centering
	\begin{subfigure}{0.24\textwidth}
		\includegraphics[width=\textwidth]{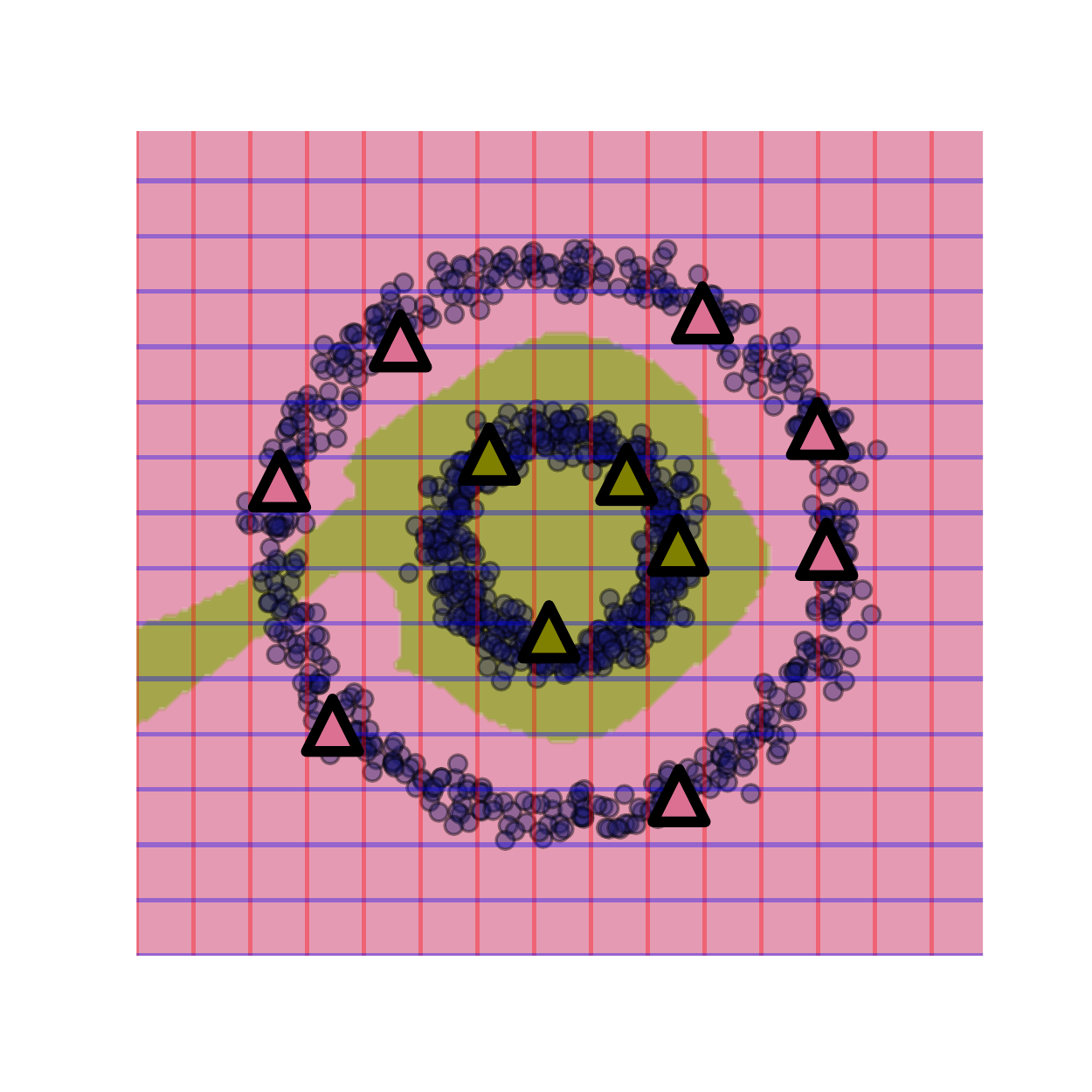}
		\caption{Labeled $+$ Unlabeled}
	\end{subfigure}
	\begin{subfigure}{0.24\textwidth}
		\includegraphics[width=\textwidth]{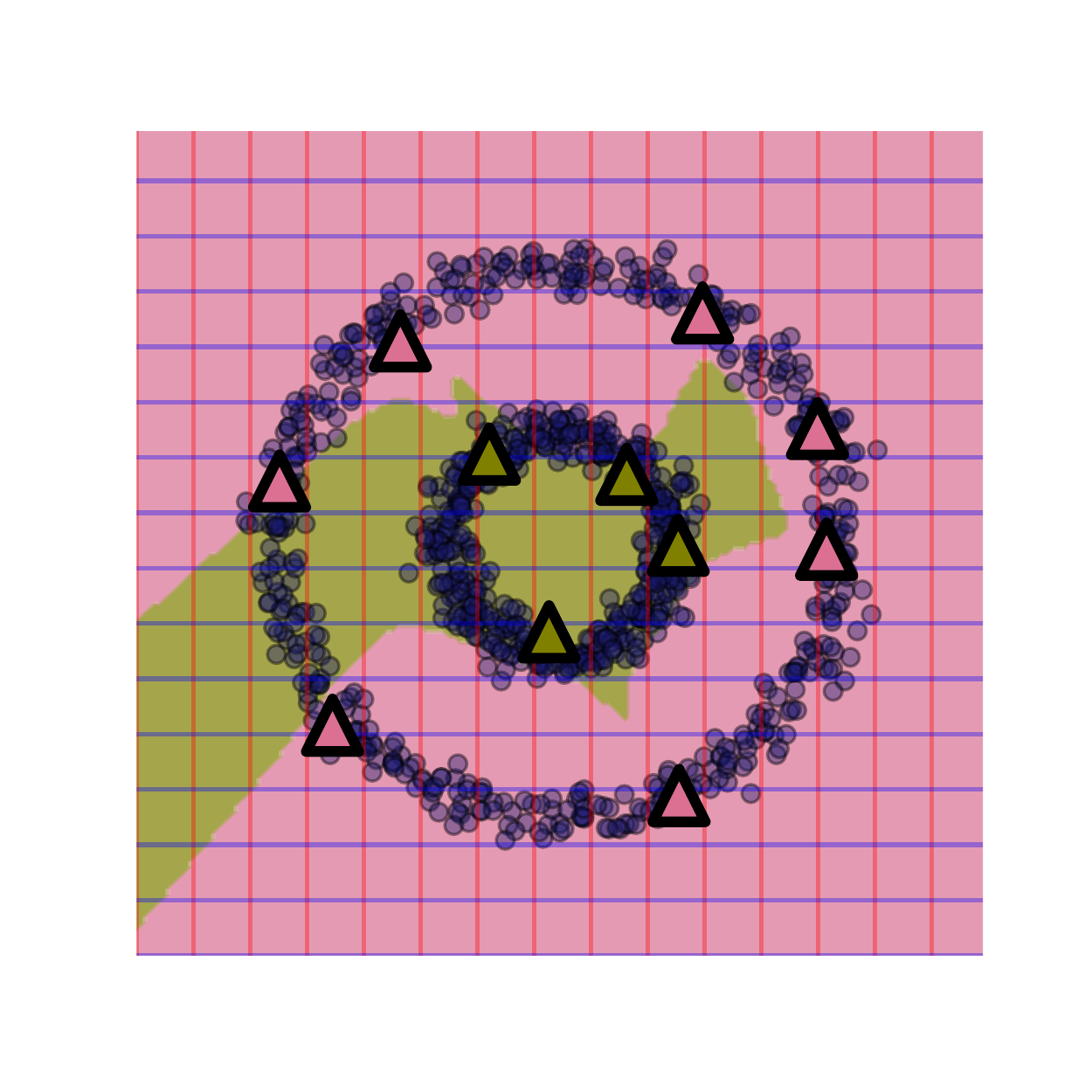}
		\caption{Labeled Only}
	\end{subfigure}
	\begin{subfigure}{0.24\textwidth}
		\includegraphics[width=\textwidth]{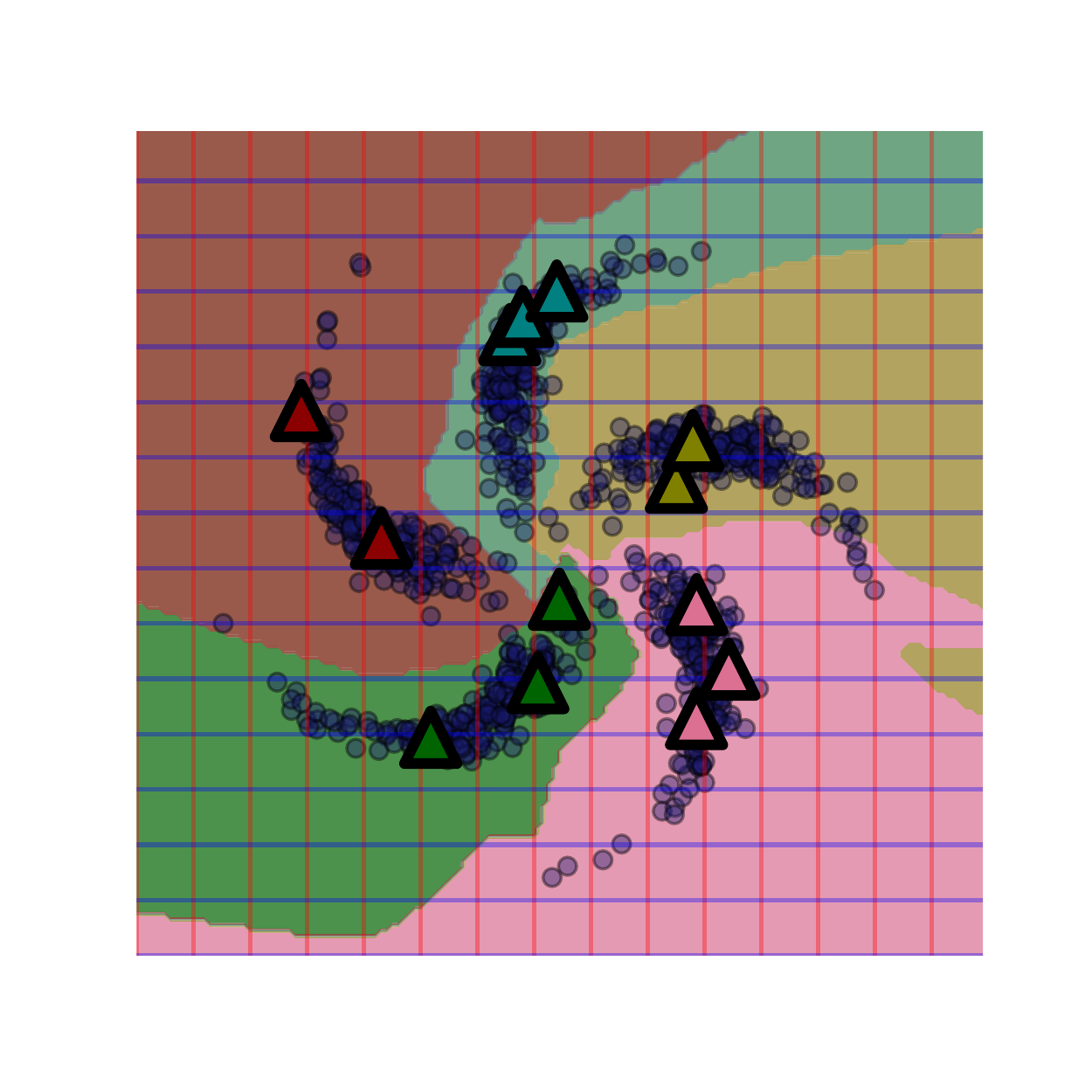}
		\caption{Labeled $+$ Unlabeled}
	\end{subfigure}
	\begin{subfigure}{0.24\textwidth}
		\includegraphics[width=\textwidth]{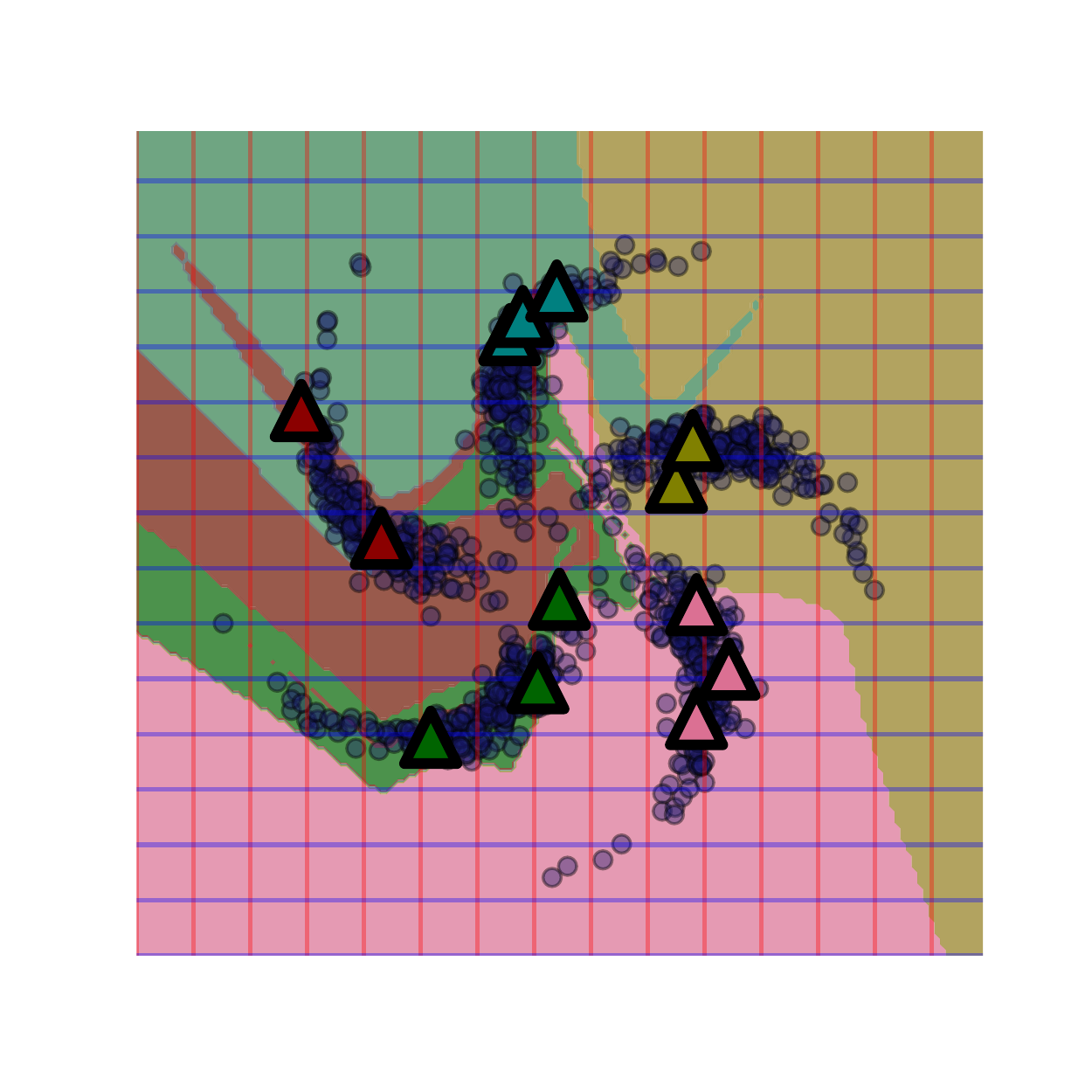}
		\caption{Labeled Only}
	\end{subfigure}
	\caption{
    Illustration of \method performance on synthetic datasets. 
    Labeled data are shown with colored triangles, and unlabeled data are shown with blue circles. 
    Colors represent different classes.
    We compare the classifier decision boundaries when only using labeled data (panels b, d) and when using both labeled and unlabeled data (panels a, c) on two circles (panels a, b) and pinwheel (panels c, d) datasets.
    \method leverages unlabeled data to push the decision boundary to low-density regions of the space.
    }
	\label{fig:toy}
    \vspace{-.5cm}
\end{figure*}

\subsection{Consistency Regularization}
\label{sec:cons}
Most of the existing state-of-the-art approaches to semi-supervised learning on image data are based 
on consistency regularization 
\citep{laine2016temporal, miyato2018virtual, tarvainen2017mean, athiwaratkun2018there, verma2019interpolation, xie2020unsupervised, 
berthelot2020remixmatch}.
These methods penalize changes in network predictions with respect to input perturbations, such as random translations and horizontal flips, with an additional loss term that can be computed on unlabeled data,
\begin{equation} \label{eq:cons_orig}
    \ell_{cons}(x) =  \| g(x') - g(x'') \|^2,
\end{equation}
 where $x', x''$ are random perturbations of $x$, and $g$ is the vector of probabilities over the classes.

Motivated by these methods, we introduce a new consistency regularization term for \method.
Let $y''$ be the label predicted on image $x''$ by \method according to \eqref{eq:preds}.
We then define our consistency loss as the negative log likelihood of the input $x'$ given the label $y''$:
\begin{multline}
    \label{eq:cons}
    L_\text{cons}(x', x'') = -\log p(x'|y'') = \\ 
    -\log \mN(f(x') \vert \mu_{y''}, \Sigma_{y''}) 
    -  \log \left| \det \left( \frac{\partial f}{\partial x'} \right) \right|.
\end{multline}
This loss term encourages the
model to map small perturbations of the same unlabeled inputs to the same 
components of the Gaussian mixture distribution in the latent space.
Unlike the standard consistency loss of \eqref{eq:cons_orig}, the proposed loss in \eqref{eq:cons} takes values on the same scale as the data log likelihood \eqref{eq:loss}, and indeed we find it to work better in practice.
We refer to \method with the consistency term as \methodcons.
The final loss for \methodcons is then the weighted sum of the consistency loss \eqref{eq:cons} and the negative log likelihood of both labeled and unlabeled data \eqref{eq:loss}.

\section{Experiments}
\label{sec:exps}

We evaluate \method on a wide range of datasets across different application domains including low-dimensional synthetic data (Section \ref{sec:exp_toy}), text and tabular data (Section \ref{sec:exp_tab}), and image data (Section \ref{sec:exp_img}). 
We show that \method outperforms the baselines on tabular and text data. \method is also state-of-the-art as an \emph{end-to-end generative} approach to semi-supervised image classification, conditioned on architecture. However, \method is constrained by the RealNVP architecture, and thus does not outperform the most powerful approaches in this setting, which involve discriminative classifiers.

In all experiments, we use the RealNVP normalizing flow architecture. Throughout training, Gaussian mixture parameters are fixed:
the means are initialized randomly from the standard normal distribution and the covariances are set to $I$. See Appendix \ref{sec:meanchoice} for further discussion on GMM initialization and training.

\subsection{Synthetic Data}
\label{sec:exp_toy}

We first apply \method to a range of two-dimensional synthetic datasets, 
in order to gain a better visual intuition for the method.
We use the RealNVP architecture with 5 coupling layers, defined 
by fully-connected shift and scale networks, each with 1 hidden 
layer of size 512.
In addition to the semi-supervised setting, we also trained the method only using the labeled data.
In Figure \ref{fig:toy} we visualize the decision boundaries of the classifier corresponding to \method for both of these settings on the \textit{two circles} and \textit{pinwheel} datasets. 
On both datasets, \method is able to benefit from the unlabeled data to push the decision boundary to a low-density region, as expected.
On the two circles problem the method is unable to fit the data perfectly as flows are homeomorphisms, and the disk is topologically distinct from an annulus.
However, \method still produces a reasonable decision boundary and improves over the case when only labeled data are available.
We provide additional visualizations in Appendix \ref{sec:synthetic},
Figure \ref{fig:all_synth}. 

\begin{table*}[t]
	\centering
	\caption{
	Accuracy on BERT embedded text classification datasets and UCI datasets with a small number of labeled examples. The kNN baseline, logistic regression, and the 3-Layer NN + Dropout were trained on the labeled data only. Numbers reported for each method are the best of $3$ runs (ranked by performance on the validation set). $n_l$ and $n_u$ are the number of labeled and unlabeled data points.
    }
	\label{tab:ssl_tab}
	\small
	\begin{tabular}{lccccc}
	    & \multicolumn{4}{c}{Dataset ($n_l$ / $n_u$, classes)}  \\
        \cmidrule(r){2-5}
	    \\[-0.3cm]
	    Method                    & AG-News  
                                  & Yahoo Answers
                                  & Hepmass 
                                  & Miniboone\\
                                  & (200 / 200k, 4)
                                  & (800 / 50k, 10)
                                  & (20 / 140k, 2)
                                  & (20 / 65k, 2)
                                  \\
                                  \midrule
        kNN & 51.3 & 28.4 & 84.6 & 77.7 \\
        Logistic Regression & 78.9 & 54.9 & 84.9 & 75.9 \\
        3-Layer NN + Dropout & 78.1 & 55.6 & 84.4 & 77.3 \\
        \midrule
        RBF Label Spreading & 54.6 & 30.4 & 87.1 & 78.8 \\
        kNN Label Spreading & 56.7 & 25.6 & 87.2 & 78.1 \\
        $\Pi$-model & 80.6 & 56.6 & 87.9 & 78.3 \\
        FlowGMM & \textbf{84.8} & \textbf{57.4} & \textbf{88.8} & \textbf{80.6} \\

	\end{tabular}
\end{table*}

\begin{table*}[t]
	\centering
	\caption{
	Accuracy of the \method, VAE model \citep[M1+M2 VAE,][]{kingma2014semi},
	DIGLM \citep{nalisnick2019hybrid} in supervised and semi-supervised settings
	on MNIST, SVHN, and CIFAR-10. 
	\method Sup (\textit{All} labels) as well as DIGLM Sup (\textit{All} labels) were trained on full train datasets with all labels to demonstrate general capacity of these models. \method Sup ($n_l$ labels) was trained on $n_l$ labeled examples (and no unlabeled data).
	For reference, at the bottom we list the performance of the $\Pi$-Model \citep{laine2016temporal} and BadGAN \citep{dai2017} as representative consistency-based and GAN-based state-of-the-art methods. Both of these methods use non-invertible architectures with substantially higher base performance and, thus, are not directly comparable. 
    }
	\label{tab:ssl}
    \small
	\begin{tabular}{lcccc}
	    & \multicolumn{3}{c}{Dataset ($n_l$ / $n_u$)}  \\
        \cmidrule(r){2-4}
	    \\[-0.3cm]
		Method                    & MNIST  
                                  & SVHN  
                                  & CIFAR-10 \\
                                  & ($1k / 59k$) 
                                  & ($1k / 72k$)
                                  & ($4k / 46k$)\\
       \midrule
        DIGLM Sup (\textit{All} labels)        & 99.27         & 95.74         & -             \\
        \method Sup (\textit{All} labels)      & 99.63         & 95.81         & 88.44         \\
		\midrule
		M1+M2 VAE SSL          & 97.60         & 63.98         & -          \\
		DIGLM SSL              & \textbf{99.0}         & -             & -          \\
        \method Sup ($n_l$ labels)         & 97.36         & 78.26         & 73.13         \\
        \method                & 98.94         & 82.42         & 78.24         \\
        \methodcons            & \textbf{99.0} & \textbf{86.44}& \textbf{80.9} \\
        \midrule
        BadGAN & - & 95.75 & 85.59\\
        $\Pi$-Model & - & 94.57 & 87.64\\
        
	\end{tabular}
	\vspace{-.3cm}
\end{table*}

\subsection{Text and Tabular Data}
\label{sec:exp_tab}

\method can be especially useful for semi-supervised learning on tabular data.
Consistency-based semi-supervised methods have mostly been developed for image classification, where the predictions of the method are regularized to be invariant to random flips and translations of the image. 
On tabular data, desirable invariances are less obvious, finding suitable transformations to apply for consistency-based methods is not-trivial.
Similarly, approaches based on GANs have mostly been developed for images. We evaluate FlowGMM on the Hepmass and Miniboone UCI classification datasets (previously used in \citet{papamakarios2017masked} for density estimation).

Along with standard tabular UCI datasets, we also consider text classification on \textrm{AG-News} and \textrm{Yahoo Answers} datasets. 
Using the recent advances in transfer learning for NLP, we construct embeddings for input texts using the BERT transformer model \citep{devlin2018bert} trained on a corpus of Wikipedia articles,
and then train \method and other baselines on the embeddings.

We compare \method to the graph based label spreading method from \citet{zhou2004learning}, a $\Pi$-Model \citep{laine2016temporal} that uses dropout perturbations, as well as supervised logistic regression, k-nearest neighbors, and a neural network trained on the labeled data only. 
We report the results in Table \ref{tab:ssl_tab}, where \method outperforms the alternative semi-supervised learning methods on each of the considered datasets. 
Implementation details for \method, the baselines, and data preprocessing details are in Appendix~\ref{sec:tabulardetails}. 

\begin{table*}[t]
	\centering
	\caption{
	Semi-supervised classification accuracy for \methodcons and VAE M1 + M2 model 
	\citep{kingma2014semi} on MNIST for different
	number of labeled data points $n_l$. 
    }
	\label{tab:labeled}
	\small
	\begin{tabular}{lccccc}
		Method                    & $n_l = 100$
		                          & $n_l = 600$  
                                  & $n_l = 1000$  
                                  & $n_l = 3000$ \\
		\midrule
		M1+M2 VAE SSL ($n_l$ labels)          & $96.67$ & $97.41 \pm 0.05$  & $97.60 \pm 0.02$        & $97.82 \pm 0.04$\\
        \methodcons ($n_l$ labels)            & $98.2$ & $98.7$ & $99$ & $99.2$ \\
	\end{tabular}
\end{table*}

\begin{table*}[t]
	\centering
	\caption{
	Negative log-likelihood and Expected Calibration Error for supervised \method trained on MNIST (1k train, 1k validation, 10k test) and CIFAR-10 (50k train, 1k validation, 9k test). \method-temp stands for tempered FlowGMM where a single scalar parameter $\sigma^2$ was learned on a validation set for variances in all components.
    }
	\label{tab:uncertainty}
	\small
	\begin{tabular}{lccccc}
 & \multicolumn{2}{c}{MNIST (test acc 97.3\%)} & \multicolumn{2}{c}{CIFAR-10 (test acc 89.3\%)}  \\
   \cmidrule(r){2-3}  \cmidrule(r){4-5}
   & \method & \method-temp & \method & \method-temp \\ \midrule
      NLL $\downarrow$   & 0.295 & 0.094 & 2.98 & 0.444 \\
      ECE $\downarrow$   & 0.024 & 0.004 & 0.108 & 0.038 \\
	\end{tabular}
\end{table*}

\subsection{Image Classification}\label{sec:exp_img}

We next evaluate the proposed method on semi-supervised image classification
benchmarks on CIFAR-10, MNIST and SVHN datasets. For all the datasets, we use the RealNVP \citep{dinh2016density} architecture. Exact implementation details are listed in the Appendix \ref{sec:imageexptsdetails}.
The supervised model is trained using the same loss \eqref{eq:loss}, where all the data points are labeled ($n_u = 0$).

We present the results for \method and \methodcons in Table \ref{tab:ssl}. 
We also report results from DIGLM \citep{nalisnick2019hybrid}, supervised only performance on MNIST and SVHN, and 
the M1+M2 VAE model \citep{kingma2014semi}. 
\method outperforms the M1+M2 model and performs better or on par with DIGLM. 
Furthermore, \methodcons improves over \method on all three datasets, 
suggesting that our proposed consistency regularization is helpful for performance.

Following \citet{oliver2018realistic}, we evaluate \methodcons varying the number of labeled data points. 
Specifically, we follow the setup of \citet{kingma2014semi} and train
\methodcons on MNIST with $100$, $600$, $1000$ and $3000$ labeled data points. We present the results in Table~\ref{tab:labeled}. \methodcons outperforms the M1+M2 model of \citet{kingma2014semi} in all the considered settings.

We note that the results presented in this Section are not directly comparable with the state-of-the-art
methods using GANs or consistency regularization
\citep[see e.g.][]{laine2016temporal,dai2017,athiwaratkun2018there, berthelot2019mixmatch}, as the architecture we employ is much less powerful for classification than the ConvNet and ResNet architectures that have been designed for classification without the constraint of invertibility. We believe that invertible architectures with better inductive biases for classification may help bridge this gap;  invertible residual networks \citep{behrmann2018invertible, chen2019residual} and invertible CNNs \citep{finzi2019invertible} are some of the early examples of this class of architectures. 

In general, it is difficult to directly compare \method with most existing approaches, because the types of architectures available for fully generative normalizing flows are very different than what is available to (partially) discriminative approaches or even other generative methods like VAEs. This difference is due to the invertibility requirement for normalizing flows.

\section{Model Analysis}
\label{sec:analysis}

We empirically analyze different aspcects of \method and highlight some useful features of this model.
In Section \ref{sec:uncertainty} we discuss the calibration of predictive uncertainties produced by the model.
In Section \ref{sec:analysis_latent}, we study the latent representations learned by \method. 
Finally, in Section \ref{sec:analysis_features}, we discuss a feature visualization technique that can be used to interpret the features learned by \method.

\begin{figure*}
\resizebox{1.0\textwidth}{!}{

\begin{tikzpicture}

\node (a) {\includegraphics[height=3.5cm]{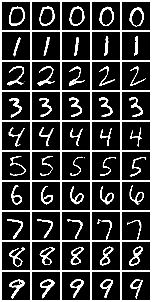}};
\node (b)[right=1.5cm] {\includegraphics[height=3.5cm]{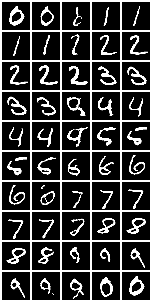}};
\node (c)[right=4.cm] {\includegraphics[height=3.7cm]{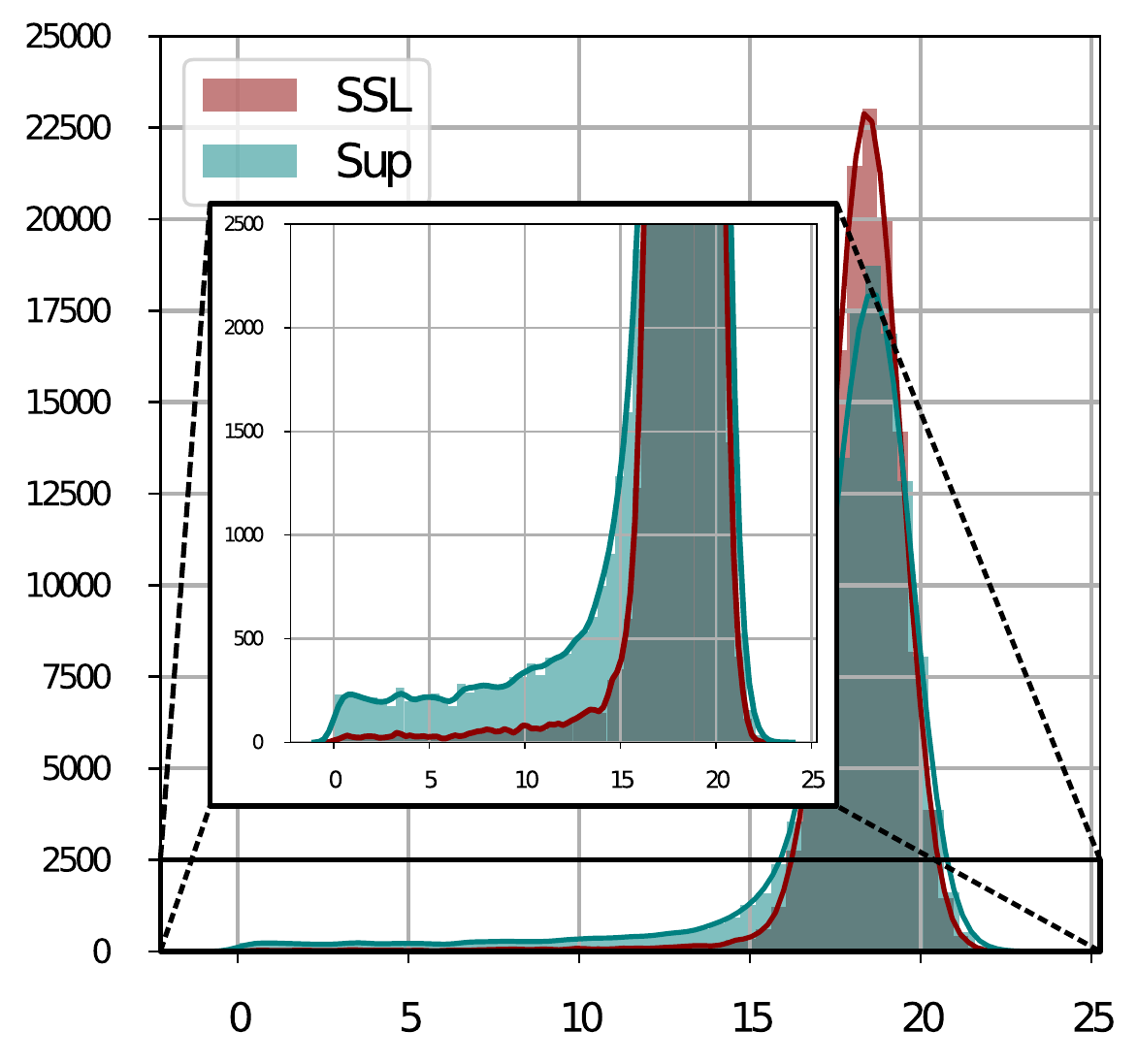}};
\node (d)[below=0.65cm, right=9.cm] {\includegraphics[width=7.cm]{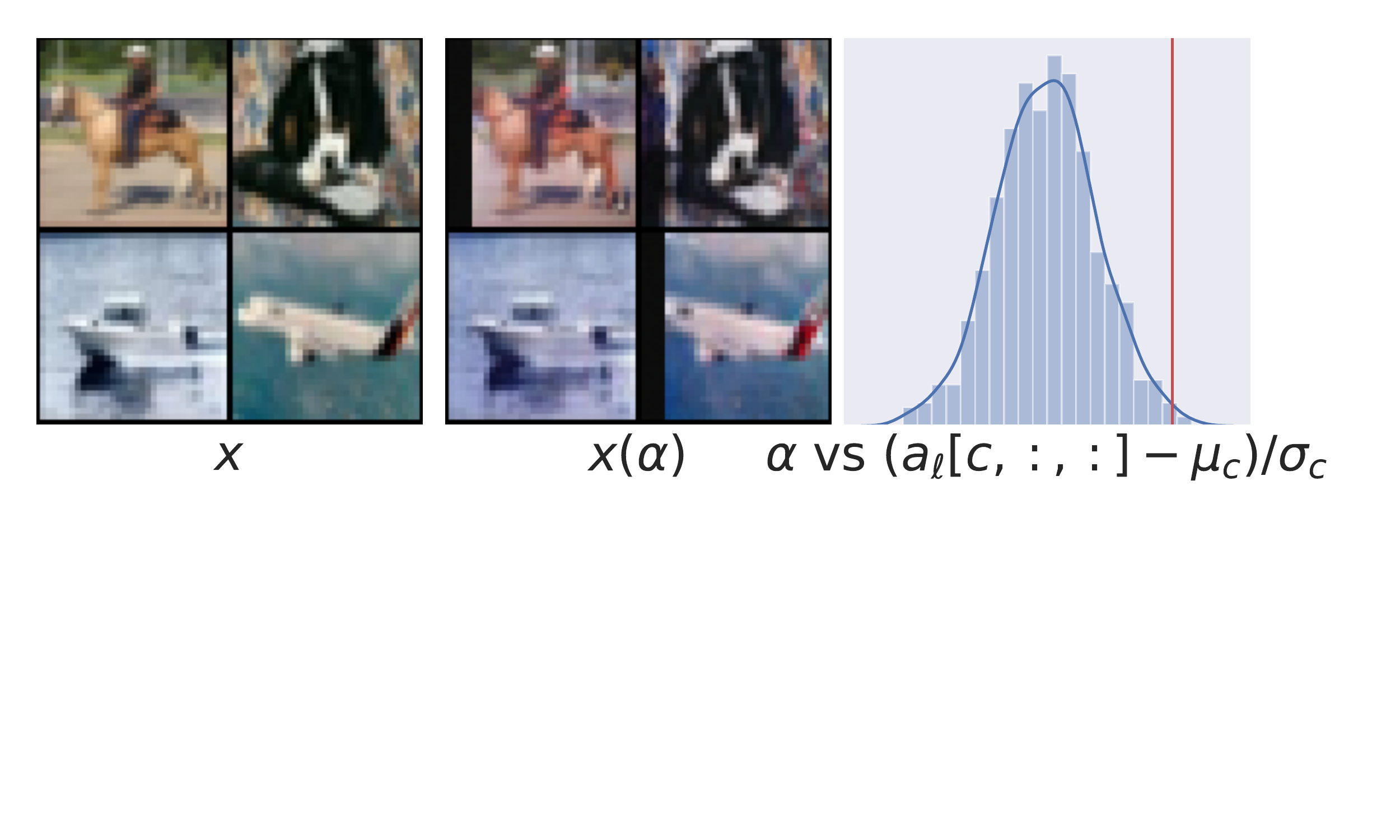}};

\draw (a.south) node[below=0mm]{(a)};
\draw (b.south) node[below=0mm]{(b)};
\draw (c.south) node[below=0mm]{(c)};
\draw (d.south) node[below=-8mm]{(d)};

\end{tikzpicture}
}
\caption{
	Visualizations of the latent space representations learned by supervised \method
	on MNIST. 
    \textbf{(a)}: Latent space interpolations between test images from the same class and
    \textbf{(b)}: from different classes. 
    Observe that interpolations between objects from different classes pass through low-density regions.
    \textbf{(c)}: Histogram of distances from unlabeled data to the decision boundary for
    \methodcons trained on $1k$ labeled and $59k$ unlabeled data and 
    \method Sup trained on $1k$ labeled data only. \methodcons is able to push the decision boundary away from the data distribution using unlabeled data.
    \textbf{(d)}: Feature visualization for CIFAR-10: four test reconstructions are shown as an intermediate feature is perturbed.  
    The value of the perturbation $\alpha$ is shown in red vs the 
    distribution of the channel activations. Observe that the channel visualized activates on zeroed out pixels to the left of the image mimicking the random translations applied to the training data.
    }
	\label{fig:latent}
    \vspace{-.3cm}
\end{figure*}

\subsection{Uncertainty and Calibration}\label{sec:uncertainty}

In many applications, particularly where decision making is involved, it is crucial to have reliable confidences associated with predictions.
In classification problems, well-calibrated models are expected to output accurate probabilities of belonging to a particular class.
 Reliable uncertainty estimation is especially relevant in semi-supervised learning since label information is limited during training. 
\citet{weinberger2017calibration}, showed that modern deep learning models are highly overconfident, but could be easily recalibrated with temperature scaling.
In this Section, we analyze the predictive uncertainties produced by \method.
In Appendix Section \ref{sec:ood}, we also consider out-of-domain data detection.

When using \method for classification, the class predictive probabilities are 
\begin{equation*}
    p(y|x) = 
    \frac{\mN \left(f(x) | \mu_{y}, \Sigma_{y}\right)}{\sum_{k=1}^{\mC} \mN(f(x) \vert \mu_k, \Sigma_k)}.
\end{equation*}
Since we initialize Gaussian mixture means randomly from the standard normal distribution and do not train them along with the flow parameters (see Appendix \ref{sec:meanchoice}), \method predictions become inherently overconfident due to the curse of dimensionality.
For example, consider two Gaussians with means sampled independently from the standard normal $\mu_1, \mu_2 \sim \mN(0, I)$ in $D$-dimensional space.  If $s_1 \sim \mN(\mu_1, I)$ is a sample from the first Gaussian, then its expected squared distances to both mixture means are $\mathbb{E} \left[\|s_1 - \mu_1\|^2 \right] = D$ and $\mathbb{E} \left[ \|s_1 - \mu_2\|^2 \right] = 3D$ (for a detailed derivation see Appendix Section \ref{sec:calibration_derivation}). In high dimensional spaces, such logits would lead to hard label assignment in \method ($p(y|x) \approx 1$ for exactly one class).
In fact, in the experiments we observe that \method is overconfident and performs hard label assignment: predicted class probabilities are all close to either $1$ or $0$. 

We address this problem by learning a single scalar parameter $\sigma^2$ for all components in the Gaussian mixture (the component $k$ will be $\mN(\mu_k, \sigma^2 I)$) by minimizing the negative log likelihood on a validation set. This way we can naturally re-calibrate the variance of the latent GMM.
This procedure is also equivalent to applying temperature scaling \citep{weinberger2017calibration} to logits $\log \mN(x|\mu_k, \Sigma_k)$.
We test FlowGMM calibration on MNIST and CIFAR datasets in the supervised setting.
On MNIST we restricted the training set size to $1000$ objects, since on the full dataset the model makes too few mistakes which makes evaluating calibration harder. 
In Table \ref{tab:uncertainty}, we report negative log likelihood and expected calibration error 
(ECE, see \citet{weinberger2017calibration} for  a description of this metric).
We can see that re-calibrating variances of the Gaussians in the mixture significantly improves both metrics and mitigates overconfidence. 
The effectiveness of this simple rescaling procedure suggests that the latent space distances learned by the flow model are correlated with the probabilities of belonging to a particular class: the closer a datapoint is to the mean of a Gaussian in the latent space, the more likely it belongs to the corresponding class.

\subsection{Learned Latent Representations}
\label{sec:analysis_latent}

We next analyze the latent representation space learned by \method. 
We examine latent interpolations between members of the same class
in Figure \ref{fig:latent}(a) and between different classes in Figure 
\ref{fig:latent}(b) for our MNIST \methodcons model trained with $n_\ell = 1k$ labels. 
As expected, inter-class interpolations pass through regions of low-density, leading to low quality samples but intra-class interpolations do not.
These observations suggest that, as expected, the model learns to put the decision boundary in the low-density region of the data space.

In Appendix section \ref{sec:samples}, we present images corresponding to the means of the Gaussian mixture and class-conditional samples from \method.

\paragraph{Distance to Decision Boundary}
To explicitly test this conclusion, we compute the distribution of distances from unlabeled data
to the decision boundary for \methodcons and \method Sup trained 
on labeled data only. 
In order to compute this distance exactly for an image $x$, we find the two
closest means $\mu'$, $\mu''$ to the corresponding latent variable
$z = f(x)$, and evaluate the expression
$d(x) = \frac{\big|\|\mu' - f(x)\|^2 - \|\mu'' - f(x)\|^2\big|}{2 \|\mu' - \mu''\|}$.
We visualize the distributions of the distances for the supervised 
and semi-supervised method in Figure~\ref{fig:latent}(c). 
While most of the unlabeled data are far from the decision boundary
for both methods, the supervised method puts a substantially larger
fraction of data close to the decision boundary. 
For example, the distance to the decision boundary is smaller than
$5$ for $1089$ unlabeled data points with supervised model, but 
only $143$ data points with \methodcons. This increased separation suggests that
\methodcons indeed pushes the decision 
boundary away from the data distribution as would be desired from the
clustering principle.

\subsection{Feature Visualization}
\label{sec:analysis_features}

Feature visualization has become an important tool for increasing the interpretability of neural networks in supervised learning. 
The majority of methods rely on maximizing the activations of a given neuron, channel, or layer over a parametrization of an input image with different kinds of image regularization \citep{szegedy2013intriguing,olah2017feature,mahendran2015understanding}. 
These methods, while effective, require iterative optimization too costly for real time interactive exploration.
In this Section we discuss a simple and efficient feature visualization technique that leverages the invertibility of \method. 
This technique can be used with any invertible model but is especially relevant for \method, where we can use feature visualization to gain insights into the classification decisions made by the model.  

Since our classification model uses a flow which is a sequence of invertible transformations $f(x) = f_{:L}(x) := f_L \circ f_{L-1} \circ \dots \circ f_1(x)$, intermediate activations can be inverted directly. This means that we can combine the methods of feature inversion and feature maximization directly by feeding in a set of input images, modifying intermediate activations arbitrarily, and inverting the representation. Given a set of activations in the $\ell^{th}$ layer $a_\ell[c,i,j] =f_{:\ell}(x)_{cij}$ with channels $c$ and spatial extent $i,j$, we may perturb a single neuron with 
\begin{equation}
    x(\alpha) = f_{:\ell}^{-1}(f_{:\ell}(x) + \alpha \sigma_c \delta_{c}),
\end{equation} 
where $\delta_{c}$ is a one hot vector at channel $c$; and $\sigma_c$ is the standard deviation of the activations in channel $c$ over the the training set and spatial locations. 
This procedure can be performed at real-time rates to explore the activation parametrized by $\alpha$ and the location $(c,i,j)$ 
without any optimization or hyper-parameters. We show the feature visualization for intermediate layers on CIFAR-10 test images in Figure \ref{fig:latent}(d). The channel being visualized appears to activate on the zeroed pixels from random translations as well as the green channel.
Analyzing the features learned by \method we can gain insight into the workings of the model.

\section{Discussion}

We proposed a simple and interpretable approach for end-to-end generative semi-supervised prediction with normalizing flows. While \method does not yet outperform the most powerful discriminative approaches for semi-supervised image classification \citep{athiwaratkun2018there, verma2019interpolation}, we believe it is a promising step towards making fully generative approaches more practical for semi-supervised tasks. As we develop improved invertible architectures, the performance of \method will also continue to improve.

Moreover, \method does outperform graph-based and consistency-based baselines on tabular data including semi-supervised text classification with BERT embeddings.
We believe that the results show promise for generative semi-supervised learning based on normalizing flows, especially for tabular tasks where consistency-based methods struggle. 

We view interpretability and broad applicability as a strong advantage of \method.
The access to latent space representations and the feature visualization technique discussed in Section \ref{sec:analysis} as well as the ability to sample from the model can be used to obtain insights into the performance of the model in practical applications.

\bibliography{main}
\bibliographystyle{icml2019}

\cleardoublepage
\appendix

\section{Expectation Maximization}\label{sec:em}

As an alternative to direct optimization of the likelihood \eqref{eq:loss}, we consider Expectation-Maximization algorithm (EM). EM is a popular approach for finding maximum likelihood estimates in mixture models. Suppose $X=\{x_i\}_{i=1}^n$ is the observed dataset, $T = \{t_i\}_{i=1}^n$ are corresponding unobserved latent variables (often denoting the component in mixture model) and $\theta$ is a vector of model parameters. EM algorithm consists of the two alternating steps: on E-step, we compute posterior probabilities of latent variables for each data point
$q(t_i | x_i) = P(t_i | x_i, \theta)$;
and on M-step, we fix $q$ and maximize the expected log likelihood of the data and latent variables with respect to~$\theta$:
$\mathbb{E}_q \log P(X, T | \theta) \rightarrow \max_{\theta}.$
The algorithm can be easily adapted to the semi-supervised setting where a subset of data is labeled with $\{y_i^l\}_{i=1}^{n_l}$: then, on E-step we have hard assignment to the true mixture component $q(t_i | x_i) = I[t_i = y_i^l]$ for labeled data points.

EM is applicable to fitting the transformed mixture of Gaussians.
We can perform the exact E-step for unlabeled data in the model since
\begin{align*}
    q(t | x) = \frac{p(x | t, \theta)}{p(x | \theta)} = \frac{\mN(f(x) \vert \mu_t, \Sigma_t) \cdot \left| \det \left( \frac {\partial f}{\partial x} \right) \right| }{\sum_{k=1}^{\mC} \mN(f(x) \vert \mu_k, \Sigma_k) \cdot \left| \det \left( \frac {\partial f}{\partial x} \right) \right| }\\ =  \frac{\mN(f(x) \vert \mu_t, \Sigma_t) }{\sum_{k=1}^{\mC} \mN(f(x) \vert \mu_k, \Sigma_k) }
\end{align*}
which coincides with the E-step of EM algorithm on Gaussian mixture model.
On M-step, the objective has the following form:
\begin{align*}
    \sum_{i=1}^{n_l} \log\left[ \mN( f_{\theta}(x^l_i) | \mu_{y^l_i}, \Sigma_{y^l_i}) \left|\frac {\partial f_{\theta}}{\partial x^l_i}\right| \right] + \\ \sum_{i=1}^{n_u} \mathbb{E}_{q(t_i | x^u_i, \theta)} \log\left[ \mN( f_{\theta}(x^u_i) | \mu_{t_i}, \Sigma_{t_i}) \left|\frac {\partial f_{\theta}}{\partial x^u_i}\right| \right].
\end{align*}
Since the exact solution is not tractable due to complexity of the flow model, we perform a stochastic gradient step to optimize the expected log likelihood with respect to flow parameters $\theta$.

Note that unlike regular EM algorithm for mixture models, we have Gaussian mixture parameters $\{(\mu_k, \Sigma_k)\}_{k=1}^{\mC}$ fixed in our experiments, and on M-step the update of $\theta$ induces the change of $z_i = f_{\theta}(x_i)$ latent space representations.

Using EM algorithm for optimization in the semi-supervised setting on MNIST dataset with 1000 labeled images, we obtain 98.97\% accuracy which is comparable to the result for \method with regular SGD training. However, in our experiments, we observed that on E-step, hard label assignment happens for unlabeled points ($q(t|x) \approx 1$ for one of the classes) because of the high dimensionality of the problem (see section \ref{sec:uncertainty}) which affects the M-step objective and hinders training.

\begin{table*}[t]
	\centering
	\caption{
	Tuned learning rates for 3-Layer NN + Dropout, $\Pi$-model and method on text and tabular tasks. 
	For kNN we report the number of neighbours. 
	All hyper-parameters were tuned via cross-validation.
	}
	\label{tab:ssl_learning_rates}
    \centering
    \small
	\begin{tabular}{lrrrr}
		Method Learning Rate      & AG-News  
                                  & Yahoo Answers
                                  & Hepmass 
                                  & Miniboone\\
       \midrule
        3-Layer NN + Dropout & 3e-4 & 3e-4 & 3e-4 & 3e-4 \\
        $\Pi$-model & 1e-3 & 1e-4 & 3e-3 & 1e-4 \\
        \method & 1e-4 & 1e-4 & 3e-3 & 3e-4 \\
        \midrule
        kNN & $k=4$ & $k=18$ & $k=9$ & $k=3$ \\

	\end{tabular}
\end{table*}

\section{Latent Distribution Mean and Covariance Choices}\label{sec:meanchoice}

\paragraph{Initialization} In our experiments, we draw the mean vectors $\mu_i$ of Gaussian mixture model randomly from the standard normal distribution $\mu_i \sim \mN(0, I)$, and set the covariance matrices to identity $\Sigma_i = I$ for all classes; we fixed GMM parameters throughout training.
However, one could potentially benefit from data-dependent placing of means in the latent space. We experimented with different initialization methods, in particular, initializing means using the mean point of latent representations of labeled data in each class: ${\mu_i = (1 / n_l^i) \sum_{m=1}^{n_l^i} f(x^i_m)}$ where $x^i_m$ represents labeled data points from class $i$ and $n^i_l$ is the total number of labeled points in that class. In addition, we can scale all means by a scalar value $\hat{\mu}_i = r \mu_i$ to increase or decrease distances between them. We observed that such initialization leads to much faster convergence of FlowGMM on semi-supervised classification on MNIST dataset, however, the final performance of the model was worse compared to the one with random mean placing. We hypothesize that it becomes easier for the flow model to warm up faster with data-dependent initialization because Gaussian means are closer to the initial latent representations, but afterwards the model gets stuck in a suboptimal solution. 

\paragraph{GMM training} FlowGMM would become even more flexible and expressive if we could learn Gaussian mixture parameters in a principled way. In the current setup where means are sampled from the standard normal distribution, the distances between mixture components are about $\sqrt{2D}$ where $D$ is the dimensionality of the data (see Appendix \ref{sec:calibration_derivation}). Thus, classes are quite far apart from each other in the latent space, which, as observed in Section \ref{sec:uncertainty}, leads to model miscalibration. Training GMM parameters can further increase interpretability of the learned latent space representations: we can imagine a scenario in which some of the classes are very similar or even intersecting, and it would be useful to represent it in the latent space. We could train GMM by directly optimizing likelihood \eqref{eq:loss}, or using expectation maximization (see Section \ref{sec:em}), either jointly with the flow parameters or iteratively switching between training flow parameters with the fixed GMM and training GMM with the fixed flow. In our initial experiments on semi-supervised classification on MNIST, training GMM jointly with the flow parameters did not improve performance or lead to substantial change of the latent representations. Further improvements require careful hyper-parameter choice which we leave for future work.

\section{Synthetic Experiments}
\label{sec:synthetic}

\begin{figure*}
\resizebox{1.0\textwidth}{!}{

\hspace{-0.5cm}
\begin{tikzpicture}

\node (a)[right=-2cm] {\includegraphics[width=4.cm]{figs/toy_circles_class_color.pdf}};
\node (b)[right=2.cm] {\includegraphics[width=4.cm]{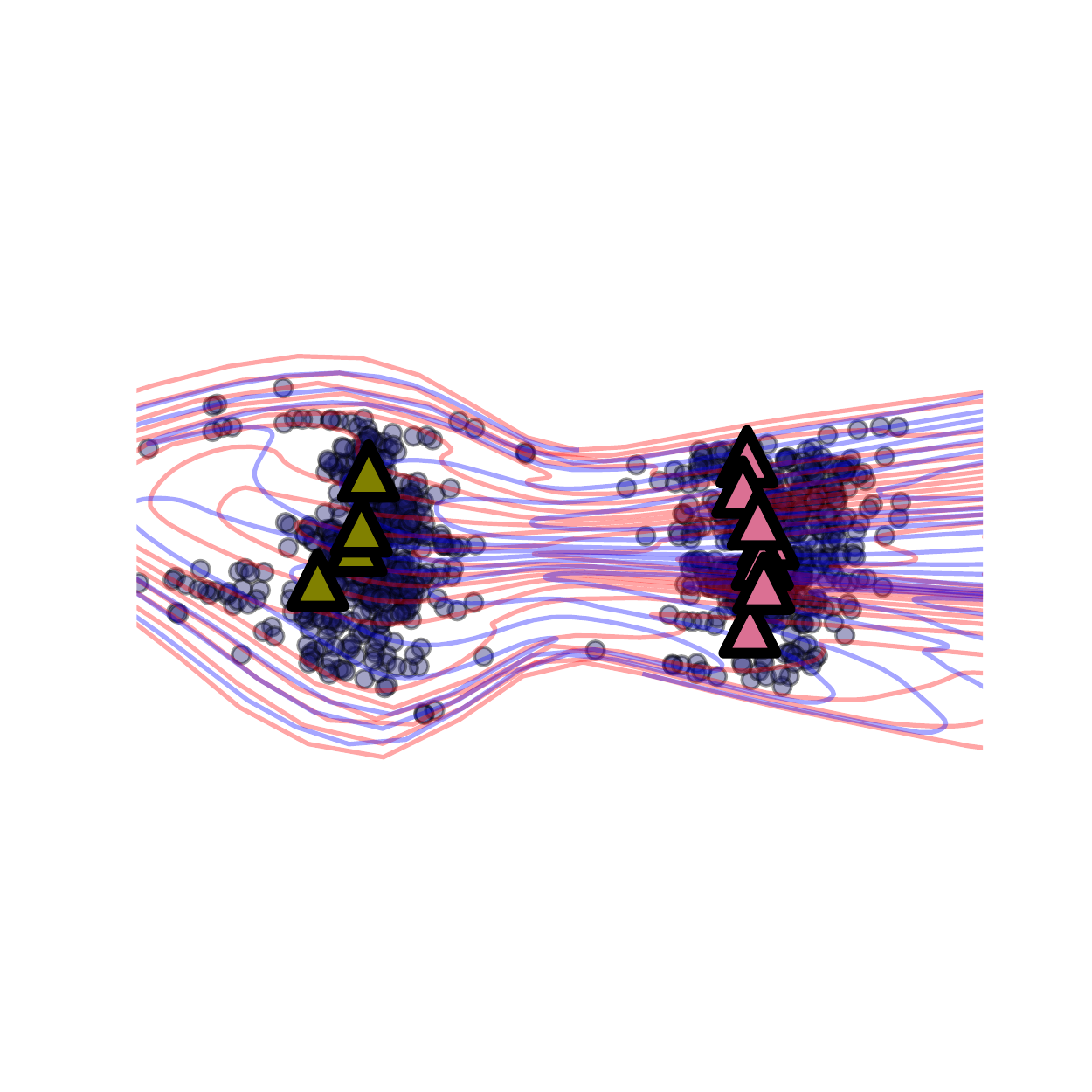}};
\node (c)[right=6.cm] {\includegraphics[width=4.cm]{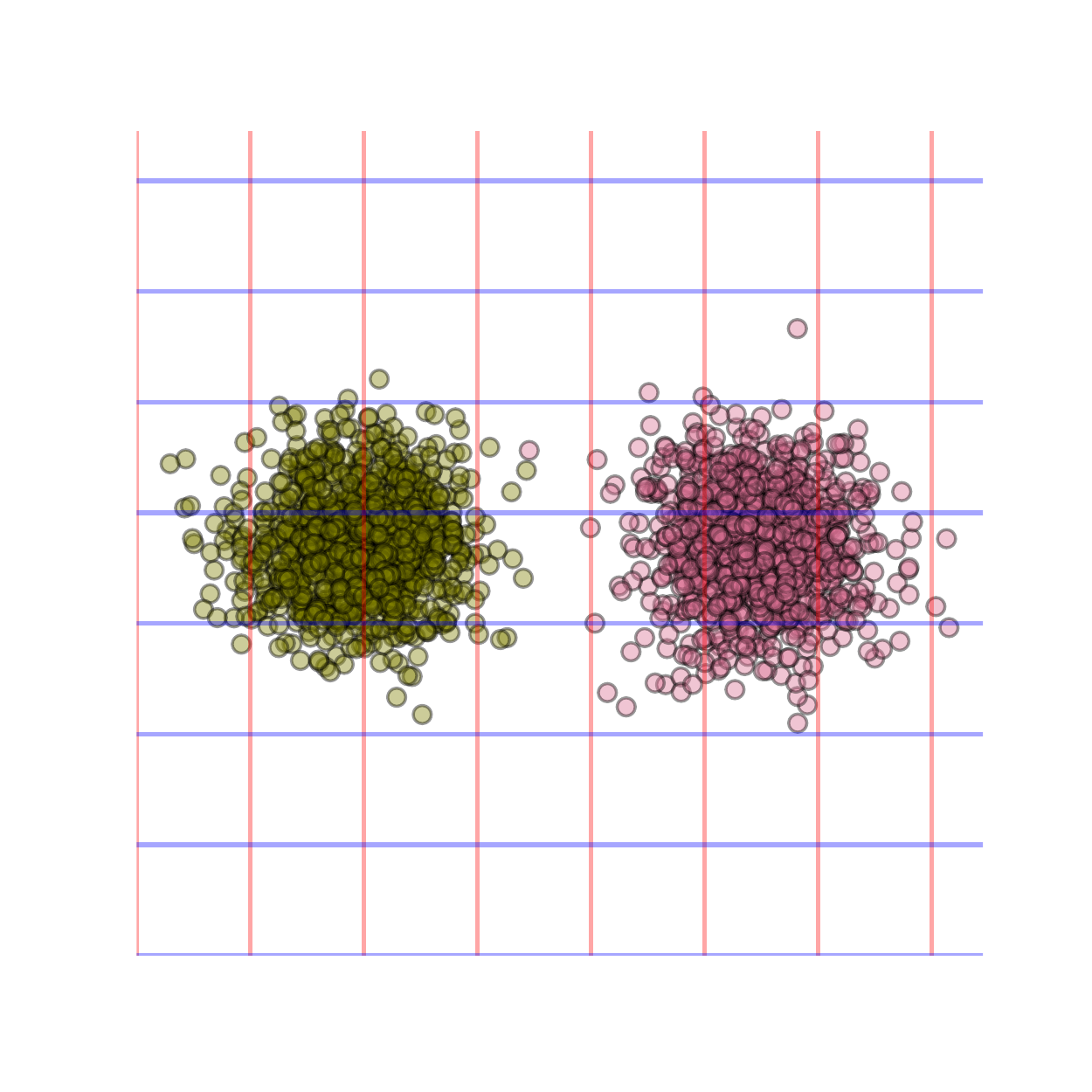}};
\node (d)[right=10.cm] {\includegraphics[width=4.cm]{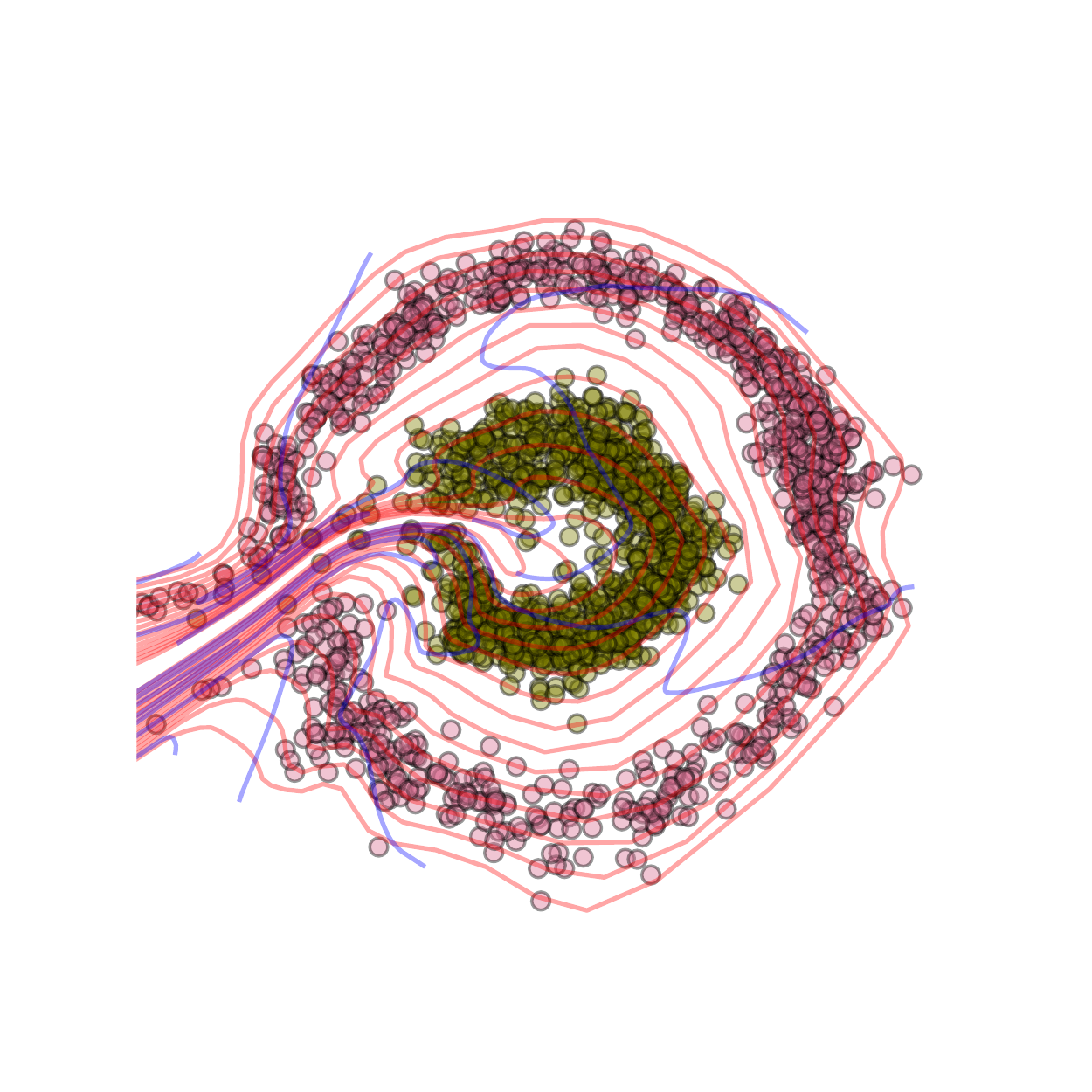}};
\node (name)[right=-2.cm, below=0cm, rotate=90] {Two Circles};

\node (a2)[below=4cm, right=-2cm]{\includegraphics[width=4.cm]{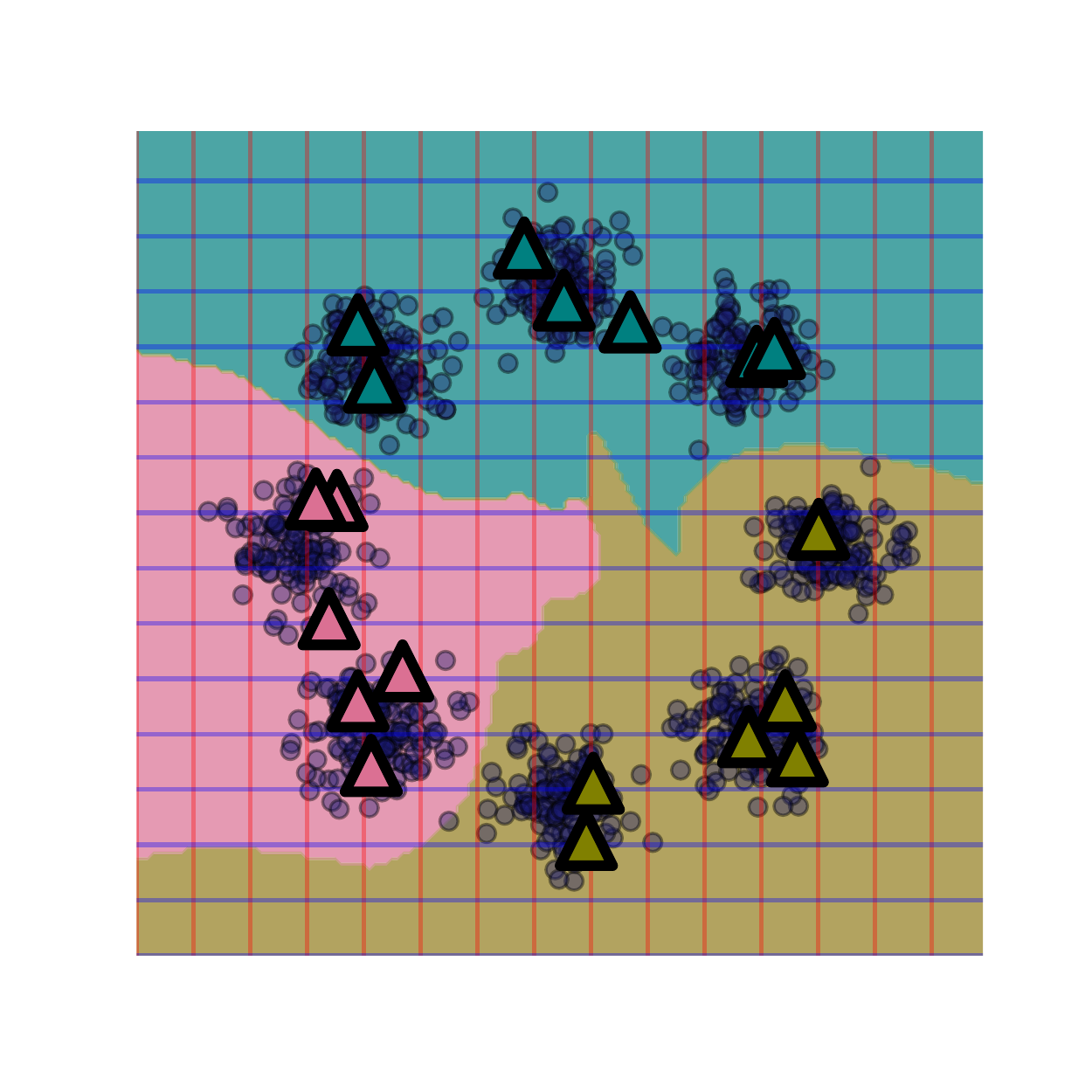}};
\node (b2)[below=4cm, right=2.cm] {\includegraphics[width=4.cm]{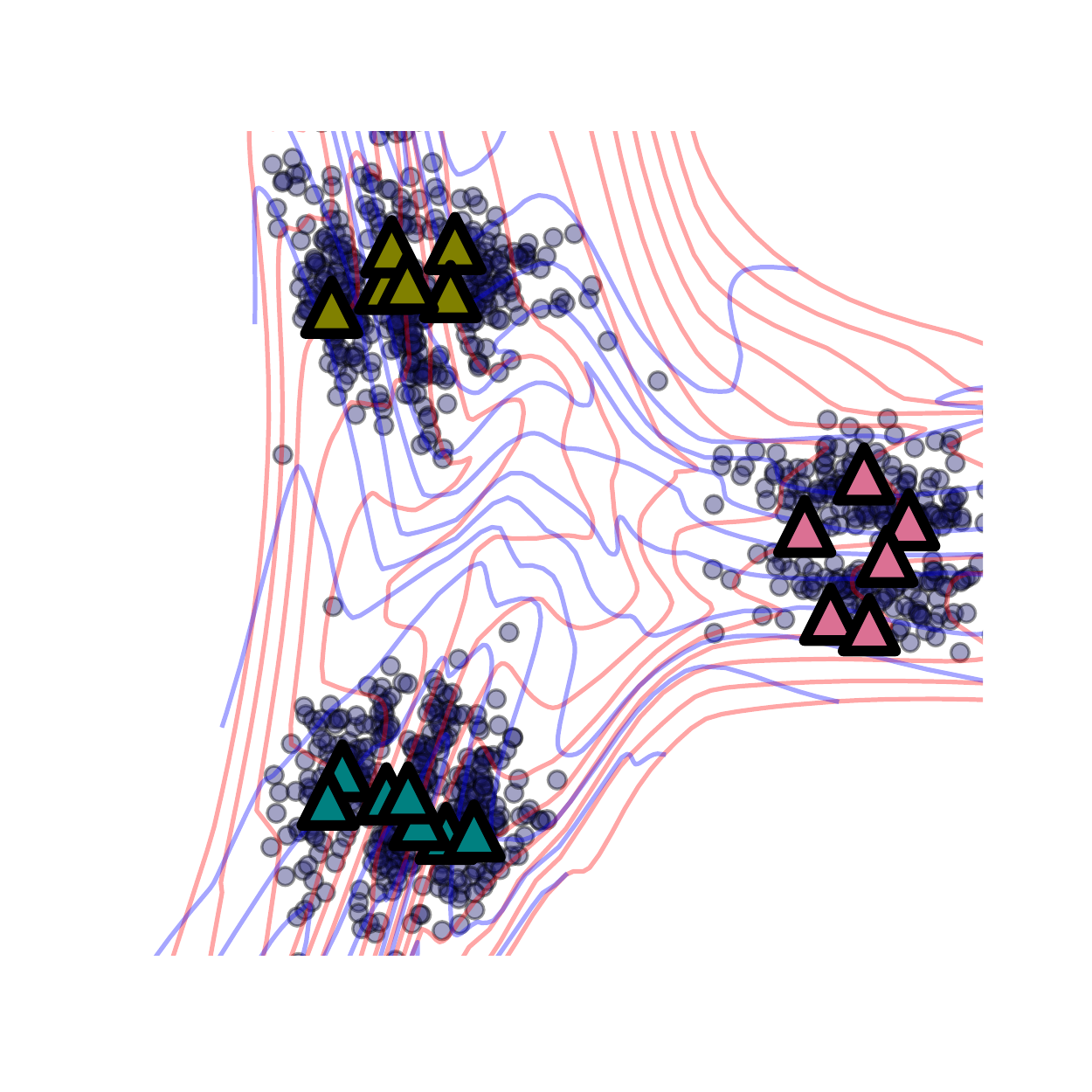}};
\node (c2)[below=4cm, right=6.cm] {\includegraphics[width=4.cm]{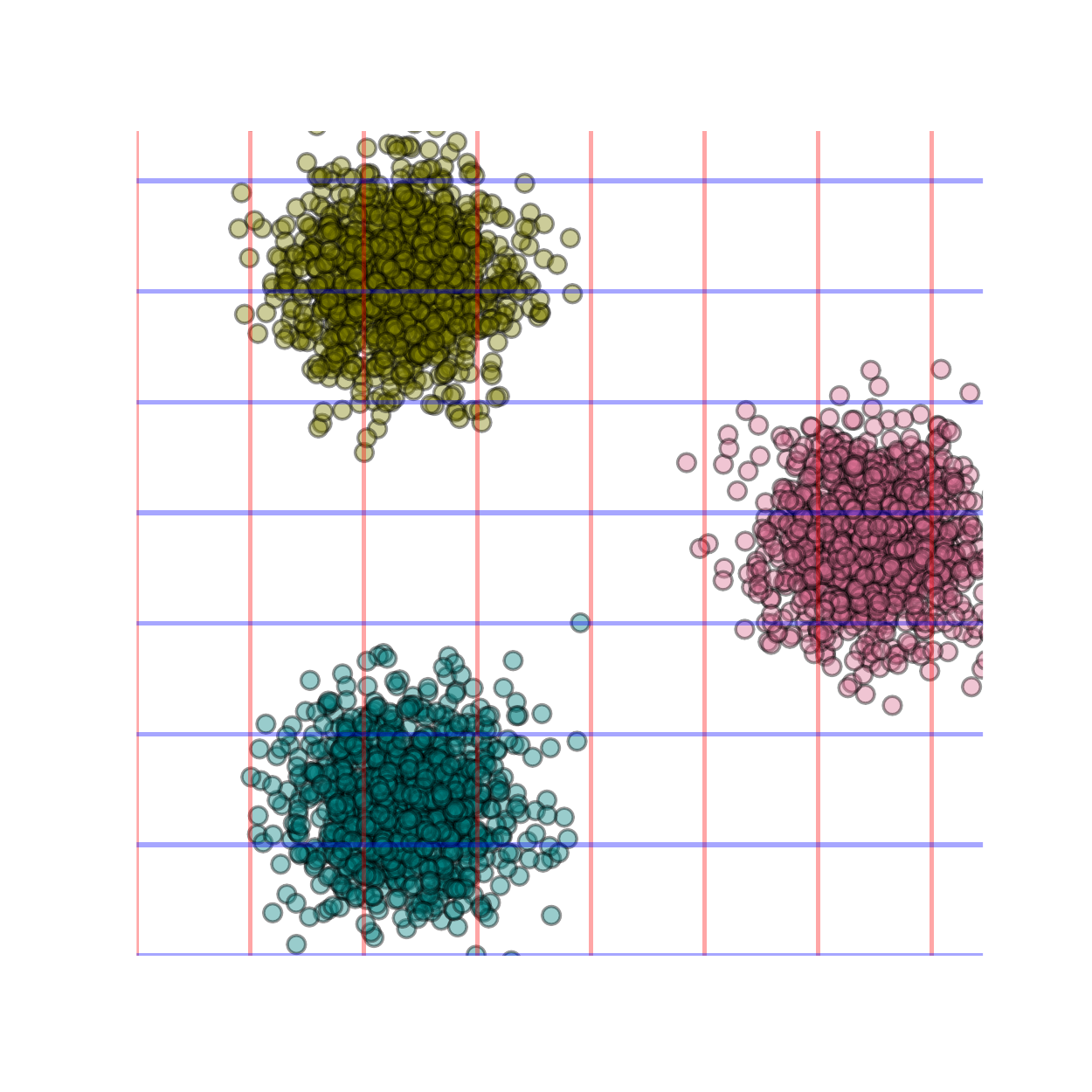}};
\node (d2)[below=4cm, right=10.cm] {\includegraphics[width=4.cm]{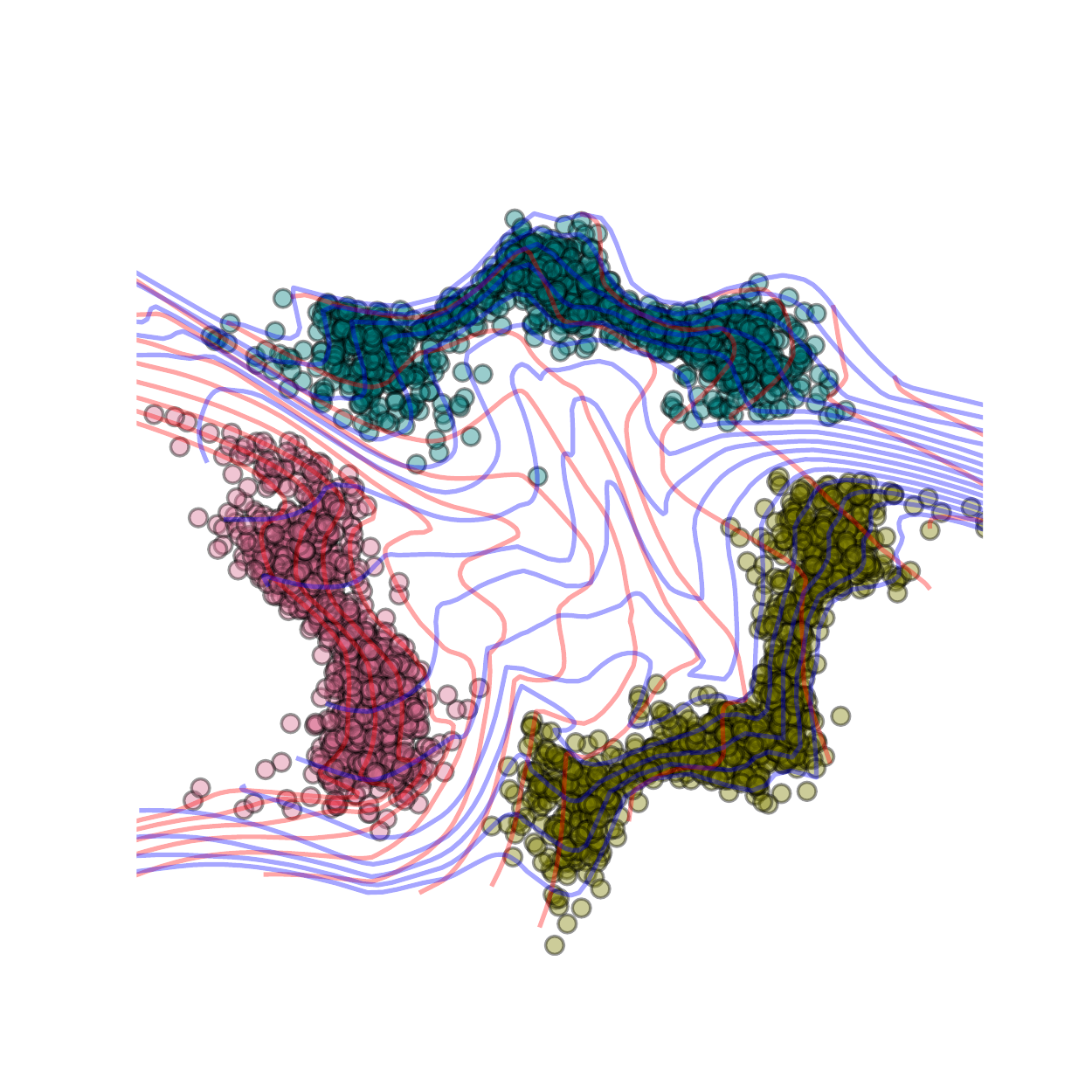}};
\node (name2)[right=-2.cm, below=4cm, rotate=90] {8 Gaussians};

\node (a3)[below=8cm, right=-2cm]{\includegraphics[width=4.cm]{figs/toy_pinwheel_class_color.pdf}};
\node (b3)[below=8cm, right=2.cm] {\includegraphics[width=4.cm]{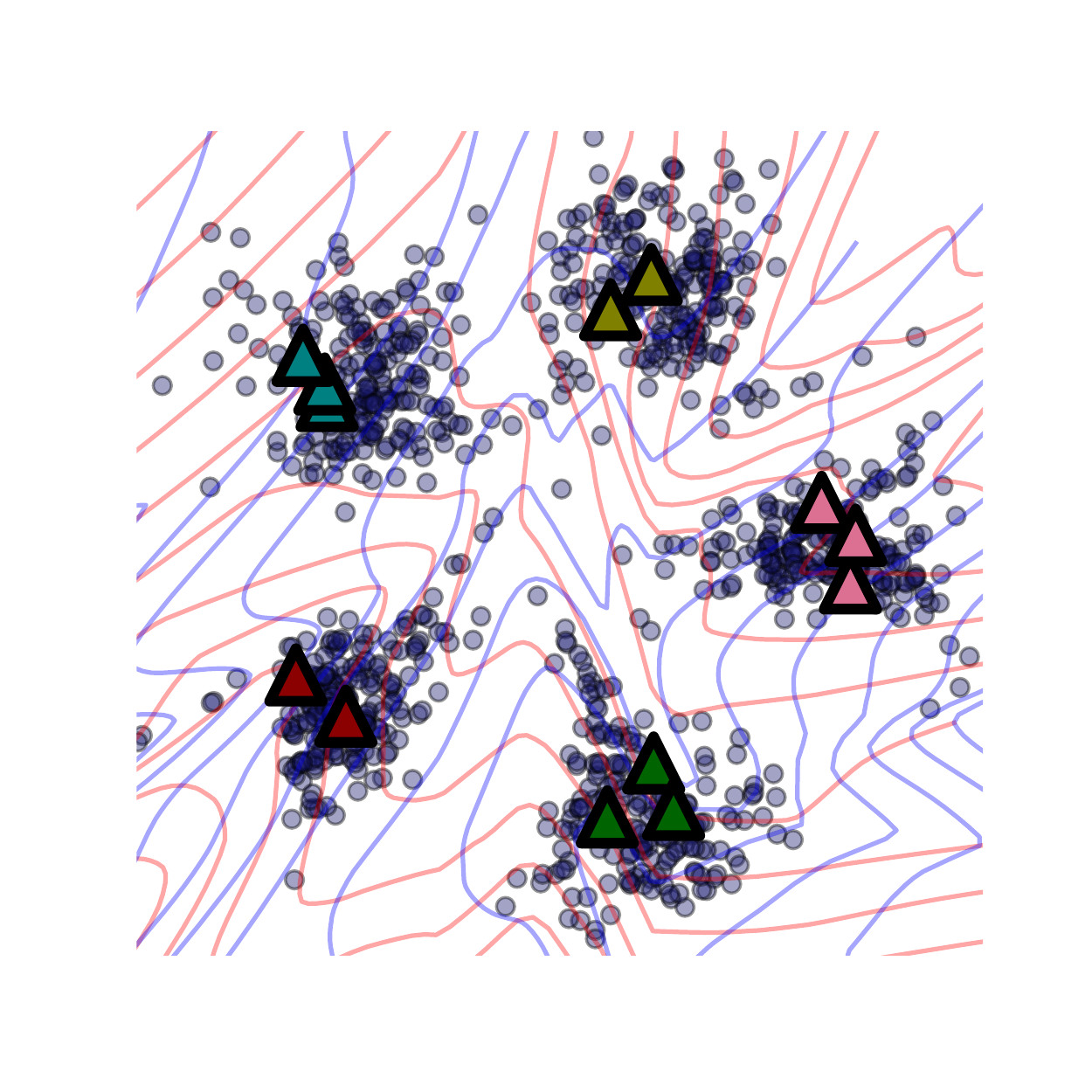}};
\node (c3)[below=8cm, right=6.cm] {\includegraphics[width=4.cm]{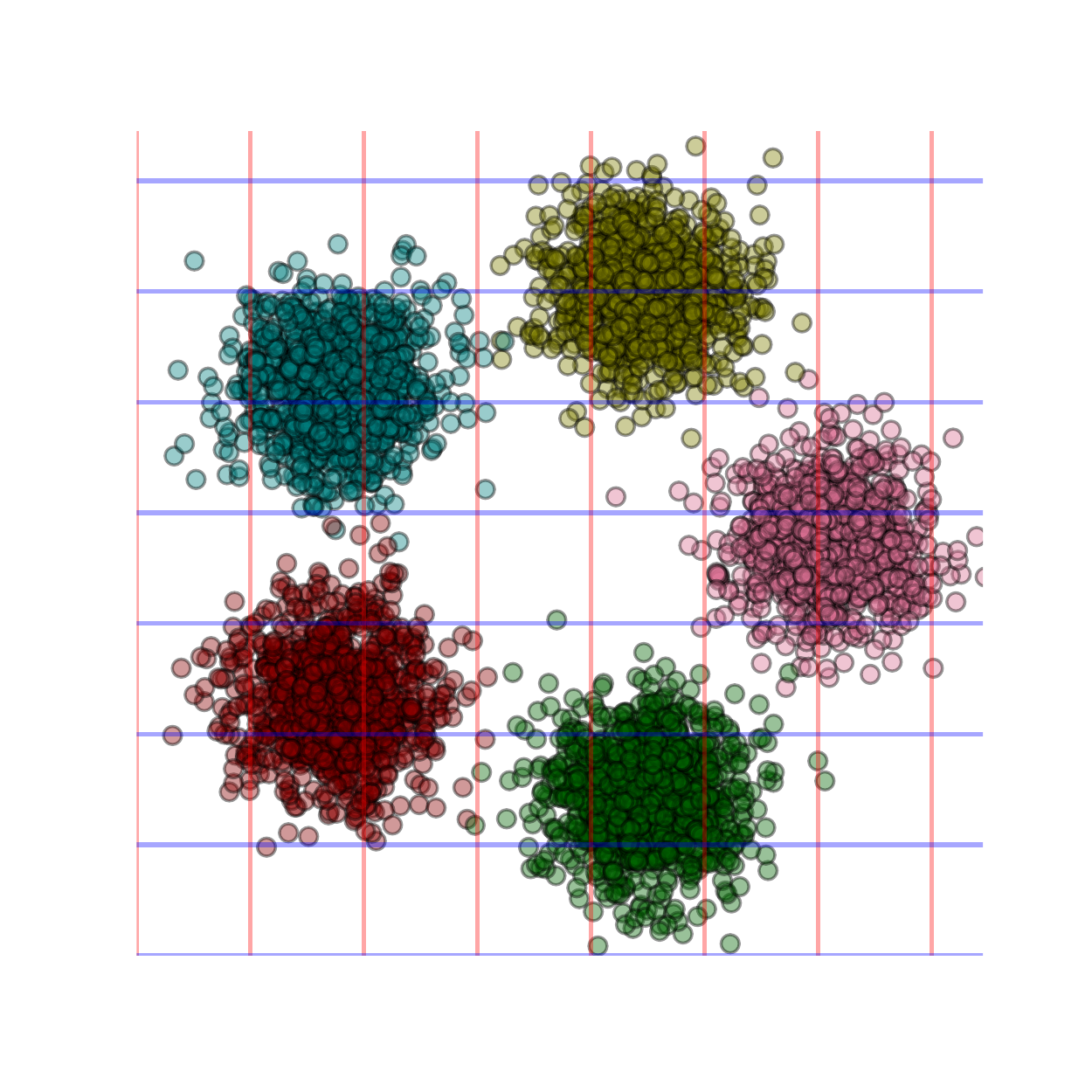}};
\node (d3)[below=8cm, right=10.cm] {\includegraphics[width=4.cm]{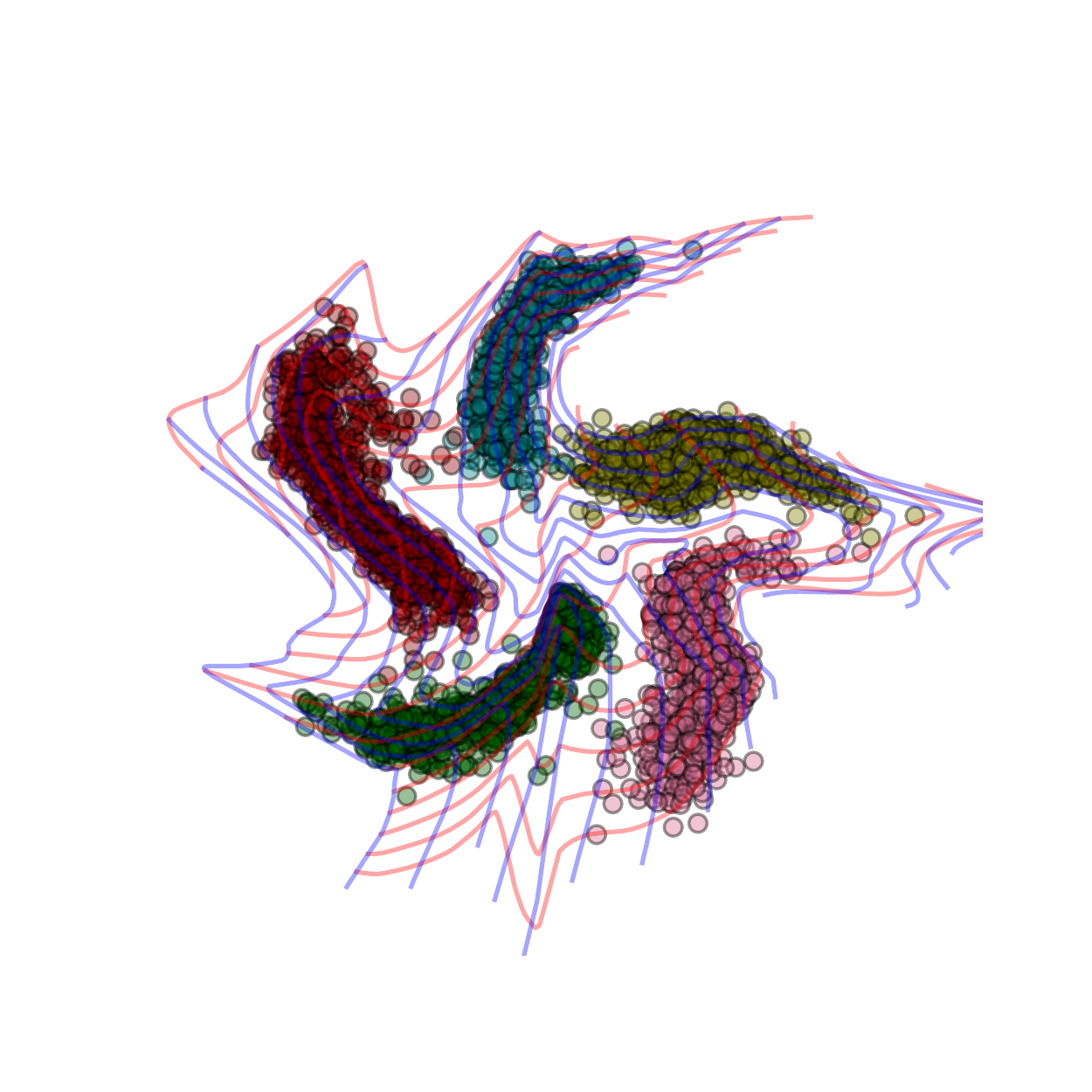}};
\node (name3)[right=-2.cm, below=8cm, rotate=90] {Pinwheel};

\draw[-latex,thick] ([xshift=-2mm] b.west) -- ([xshift=3mm] a.east) 
node[midway,below]{$f$};
\draw[-latex,thick] ([yshift=-4cm, xshift=-2mm] b.west) -- ([yshift=-4cm, xshift=3mm] a.east) 
node[midway,below]{$f$};
\draw[-latex,thick] ([yshift=-8cm, xshift=-2mm] b.west) -- ([yshift=-8cm, xshift=3mm] a.east) 
node[midway,below]{$f$};

\draw[-latex,thick] ([xshift=-1mm] d.west) -- ([xshift=3mm] c.east) 
node[midway,below]{$f^{-1}$};
\draw[-latex,thick] ([yshift=-4cm, xshift=-1mm] d.west) -- ([yshift=-4cm, xshift=3mm] c.east) 
node[midway,below]{$f^{-1}$};
\draw[-latex,thick] ([yshift=-8cm, xshift=-1mm] d.west) -- ([yshift=-8cm, xshift=3mm] c.east) 
node[midway,below]{$f^{-1}$};

\draw (a.north) node[above=-5mm]{$\mX$, Data};
\draw (b.north) node[above=-5mm]{$\mZ$, Latent};
\draw (c.north) node[above=-5mm]{$\mZ$, Latent};
\draw (d.north) node[above=-5mm]{$\mX$, Data};
\draw (a.south) node[below=8cm]{(a)};
\draw (b.south) node[below=8cm]{(b)};
\draw (c.south) node[below=8cm]{(c)};
\draw (d.south) node[below=8cm]{(d)};

\end{tikzpicture}
}
\caption{
        Illustration of \method on synthetic datasets: 
        two circles (top row), eight Gaussians (middle row) and pinwheel (bottom row).
        \textbf{(a):} Data distribution and classification decision boundaries. Unlabeled data are shown with blue circles and labeled data are shown with colored triangles, where color represents the class. Background color visualizes the classification decision boundaries of \method.
        \textbf{(b):} Mapping of the data to the latent space.
        \textbf{(c):} Gaussian mixture in the latent space.
        \textbf{(d):} Samples from the learned generative model corresponding to different classes, as shown by their color.
    }
	\label{fig:all_synth}
    \vspace{-.5cm}
\end{figure*}

In Figure \ref{fig:all_synth} we visualize the classification decision boundaries of \method as well as the learned mapping to the latent space and generated samples for three different synthetic datasets. 

\section{Tabular data preparation and hyperparameters}\label{sec:tabulardetails}
The AG-News and Yahoo Answers were constructed by applying BERT embeddings to the text input, yielding a $768$ dimensional vector for each data point. AG-News has $4$ classes while Yahoo Answers has $10$. The UCI datasets Hepmass and Miniboone were constructed using the data preprocessing from \citet{papamakarios2017masked}, but with the inclusion of the removed background process class so that the two problems can be used for binary classification. We then subsample the fraction of background class examples so that the dataset is balanced. For each of the datasets, a separate validation set of size $5$k was used to tune hyperparameters. All neural network models use the ADAM optimizer \citep{kingma2014adam}. 

\textbf{k-Nearest Neighbors}: We tested both using both L2 distance and L2 with inputs normalized to unit norm, ($\sin^2$ distance), and the latter performed the best. The value $k$ chosen in the method was found sweeping over $1-20$, and the optimal values for each of the datasets are shown in \ref{tab:ssl_learning_rates}.

\textbf{3 Layer NN + Dropout}: The $3$-Layer NN + Dropout baseline network has three fully connected hidden layers with inner dimension $k=512$, \textrm{ReLU} nonlinearities, and dropout with $p=0.5$. We use the learning rate $3$e$-4$ for training the supervised baseline across all datasets.

\textbf{$\Pi$-Model}: The $\Pi$-Model uses the same network architecture, and dropout for the perturbations. The additional consistency loss per unlabeled data point is computed as $L_\textrm{Unlab} = ||g(x'')-g(x')||^2$, where $g$ is are the output probabilities after the softmax layer of the neural network and the consistency weight $\lambda = 30$ which worked the best across the datasets. The model was trained for $50$ epochs with labeled and unlabeled batch size $n_\ell$ for AG-News and Yahoo Answers, and labeled and unlabeled batch sizes $n_\ell$ and $2000$ for Hepmass and Miniboone.

\textbf{Label Spreading}: We use the local and global consistency method from \citet{zhou2004learning}, $Y^* = (I-\alpha S)^{-1}Y$ where in our case $Y$ is the matrix of labels for the labeled, unlabeled, and test data but filled with zeros for unlabeled and test. $S = D^{-1/2}WD^{-1/2}$ computed from the affinity matrix $W_{ij} = \exp{(-\gamma \sin^2(x_i,x_j))}$ where $\sin^2(x_i,x_j) := 1 - \frac{\langle x_i,x_j\rangle}{\|x_i\|\|x_j\|}$. This is equivalent to L2 distance on the inputs normalized to unit magnitude. Because the algorithm scales poorly with number of unlabeled points for dense affinity matrices, $O(n_u^3)$, we we subsampled the number of unlabeled data points to $10k$ and test data points to $5k$ for this graph method. However, we also evaluate the label spreading algorithm with a sparse kNN affinity matrix on using a larger subset $20k$ of unlabeled data. The two hyperparameters for label spreading ($\gamma$/$k$ and $\alpha$) were tuned by separate grid search for each of the datasets. In both cases, we use the inductive variant of the algorithm where the test data is not included in the unlabeled data. 

\textbf{\method}:
We train our \method model with a RealNVP normalizing flow, similar to the architectures used in \citet{papamakarios2017masked}. Specifically, the model uses $7$ coupling layers, with $1$ hidden layer each and $256$ hidden units for the UCI datasets but $512$ for text classification. UCI models were trained for $50$ epochs of unlabeled data and the text datasets were trained for $30$ epochs of unlabeled data. The labeled and unlabeled batch sizes are the same as in the $\Pi$-Model.

The tuned learning rates for each of the models that we used for these experiments are shown in Table \ref{tab:ssl_learning_rates}.

\begin{figure*}[t]
	\includegraphics[width=0.3\linewidth]{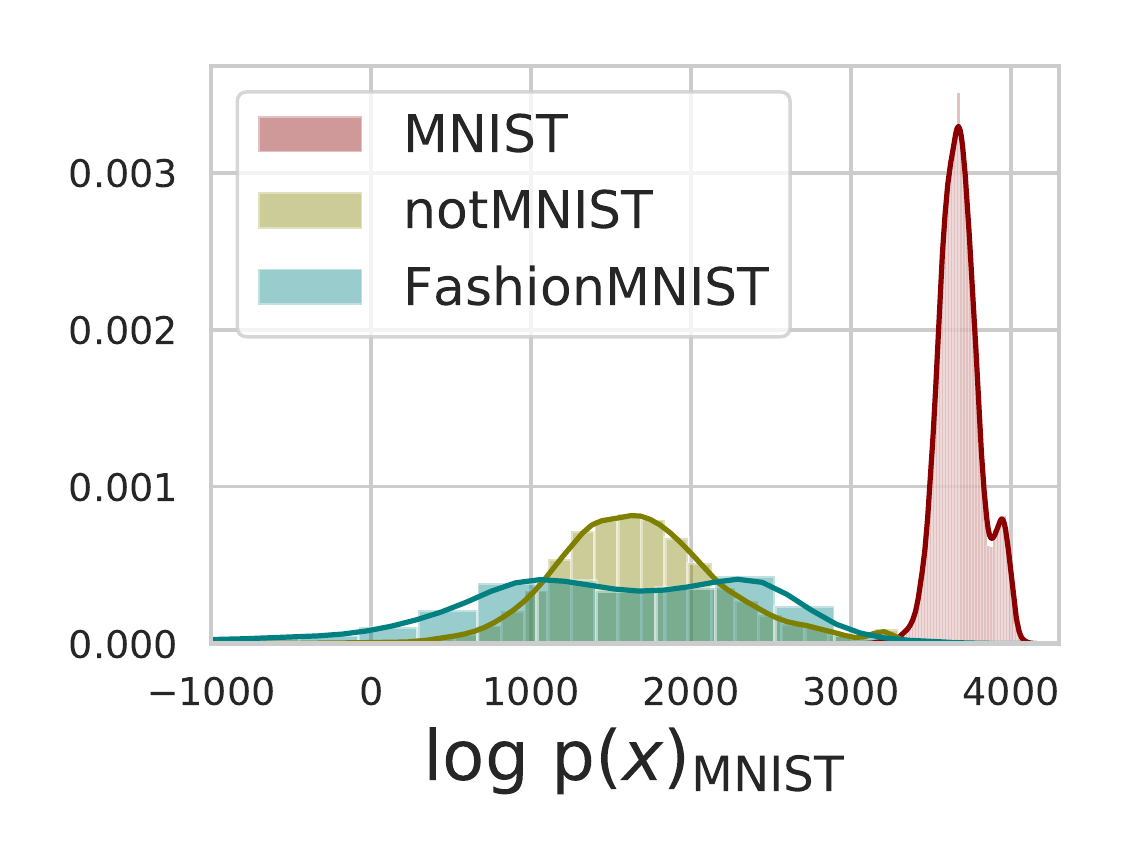}
	\includegraphics[width=0.3\linewidth]{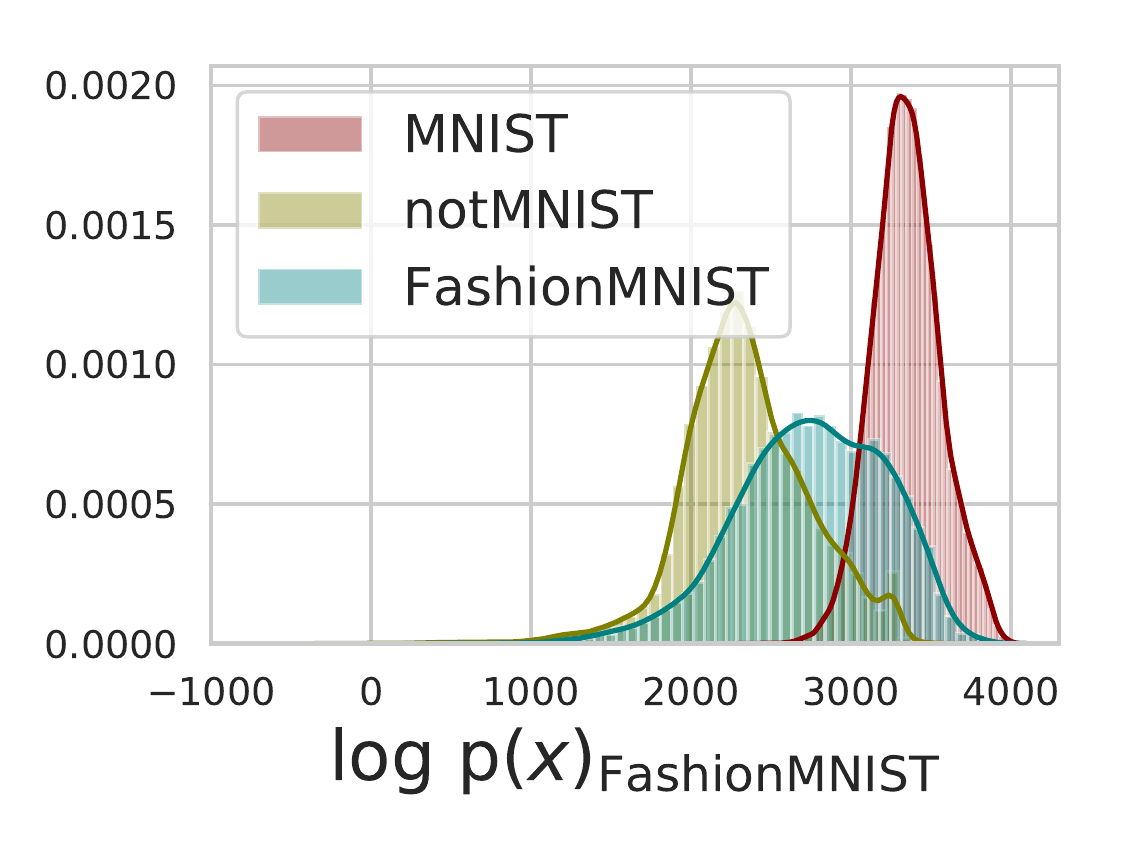}
	\includegraphics[width=0.35\linewidth]{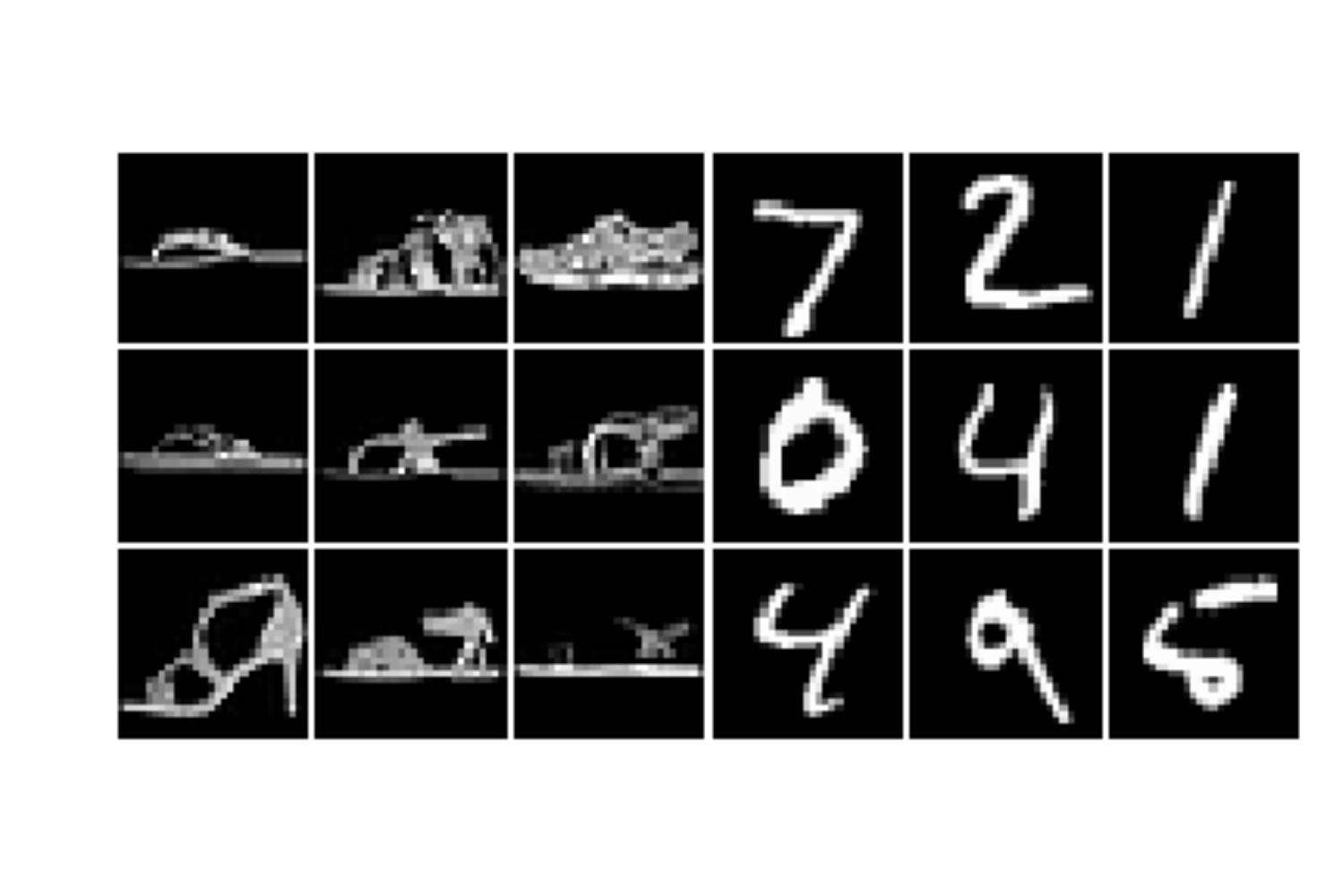}
	\caption{
	\textbf{Left:} Log likelihoods on in- and out-of-domain data for our model trained on MNIST.
    \textbf{Center:} Log likelihoods on in- and out-of-domain data for our model trained on FashionMNIST.
    \textbf{Right:} MNIST digits get mapped onto the sandal mode of the FashionMNIST model $75\%$ of the time, often being assigned higher likelihood than elements of the original sandal class. Representative elements are shown above.
	}
	\label{fig:mnist_notmnist}
    \vspace{-.3cm}
\end{figure*}

\section{Image data preparation and hyperparameters}\label{sec:imageexptsdetails}
We use the CIFAR-10
multi-scale architecture with $2$ scales, each containing $3$ coupling layers
defined by $8$ residual blocks with $64$ feature maps. We use Adam optimizer
\citep{kingma2014adam} with learning rate $10^{-3}$ for CIFAR-10 and SVHN
and $10^{-4}$ for MNIST. We train the supervised model for $100$ epochs,
and semi-supervised models for $1000$ passes through the labeled data for CIFAR-10 and SVHN and $3000$ passes for MNIST. We use a batch
size of $64$ and sample $32$ labeled and $32$ unlabeled data points in each 
mini-batch.
For the consistency
loss term \eqref{eq:cons}, we linearly increase the weight from~$0$ to~$1$
for the first $100$ epochs following \citet{athiwaratkun2018there}. For \method and \methodcons, we re-weight the loss on labeled data 
by $\lambda = 3$ (value tuned on validation \citep{kingma2014semi} on CIFAR-10), as otherwise, we observed that the method underfits the labeled data. 

\section{Out-of-domain data detection}

Density models have held promise for being able to detect out-of-domain data, an especially important task for robust machine learning systems \citep{nalisnick2019hybrid}. Recently, it has been shown that existing flow and autoregressive density models are not as apt at this task as previously thought, yielding high likelihood on images coming from other (simpler) distributions. The conclusion put forward is that datasets like SVHN are encompassed by, or have roughly the same mean but lower variance than, more complex datasets like CIFAR-10 \citep{nalisnick2018deep}.
We examine this hypothesis in the context of our flow model which has a multi-modal latent space distribution unlike methods considered in \citet{nalisnick2018deep}.
\label{sec:ood}

Using a fully supervised model trained on MNIST, we evaluate the log likelihood for data points coming from the NotMNIST dataset, consisting of letters instead of digits, and the FashionMNIST dataset. We then train a supervised model on the more complex dataset FashionMNIST and evaluate on MNIST and NotMNIST.
The distribution of the log likelihood $\log p_\mX(\cdot) = \log p_\mZ(f(\cdot)) + \log \left| \det \left( \frac {\partial f}{\partial x} \right) \right|$ on these datasets is shown in Figure \ref{fig:mnist_notmnist}. For the model trained on MNIST we see that the data
from Fashion MNIST and NotMNIST is assigned lower
likelihood, as expected.
However, the model trained on FashionMNIST predicts higher likelihoods for MNIST images. The majority ($\approx 75\%$) of the MNIST data points get mapped into the mode of the FashionMNIST model corresponding to sandals, which is the class with the largest fraction of pixels that are zero. 
Similarly, for the model trained on MNIST the image 
of all zeros has very high likelihood and gets mapped to the mode corresponding to the digit $1$ which has 
the largest fraction of empty space.

\section{Expected Distances between Gaussian Samples}
\label{sec:calibration_derivation}

Consider two Gaussians with means sampled independently from the standard normal ${\mu_1, \mu_2 \sim \mN(0, I)}$ in $D$-dimensional space.  If ${s_1 \sim \mN(\mu_1, I)}$ is a sample from the first Gaussian, then its expected squared distances to both mixture means are:
\begin{multline*}
    \mathbb{E} \left[\|s_1 - \mu_1\|^2 \right]
= \mathbb{E} \left[ \mathbb{E} \left[\|s_1 - \mu_1\|^2 | \mu_1 \right] \right]\\
= \mathbb{E} \left[ \sum_{i=1}^D \mathbb{E}\left[(s_{1, i} - \mu_{1, i})^2 | \mu_{1, i}\right] \right]\\ 
= \mathbb{E} \left[ \sum_{i=1}^D \left( \mathbb{E}[s_{1, i}^2] - 2 \mu_{1, i}^2 + \mu_{1, i}^2 \right) \right]\\
= \mathbb{E} \left[ \sum_{i=1}^D \left( 1 + \mu_{1, i}^2 - \mu_{1, i}^2 \right) \right] = D
\end{multline*}

\begin{multline*}
    \mathbb{E} \left[ \|s_1 - \mu_2\|^2 \right]
= \mathbb{E} \left[ \mathbb{E} \left[\|s_1 - \mu_2\|^2 | \mu_1, \mu_2 \right] \right] \\
= \mathbb{E} \left[ \sum_{i=1}^D \mathbb{E}\left[(s_{1, i} - \mu_{2, i})^2 | \mu_{1, i}, \mu_{2, i}\right] \right] \\
= \mathbb{E} \left[ \sum_{i=1}^D \left( 1 + \mu_{1, i}^2 - 2 \mu_{1, i}\mu_{2, i} + \mu_{2, i}^2 \right) \right] = 3D 
\end{multline*}

For high-dimensional Gaussians the random variables $\|s_1- \mu_1\|^2$ and $\|s_1 - \mu_2\|^2$ will be concentrated around their expectations.
Since the function $\exp(-x)$ decreases rapidly to zero for positive $x$, the probability of $s_1$ belonging to the first Gaussian $\exp(-\|s_1 - \mu_1\|^2) / \left( \exp(-\|s_1 - \mu_1\|^2) + \exp(-\|s_1 - \mu_2\|^2) \right) \approx \exp(-D) / (\exp(-D) + \exp(-3 D)) = 1 / (1 + \exp(-2D))$ saturates at 1 with the growth of dimensionality $D$.

\section{\method as generative model}
\label{sec:samples}

\begin{figure}[h!]
	\centering
	\begin{subfigure}{0.02\textwidth}
		\includegraphics[height=10\textwidth]{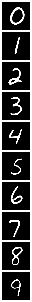}
		\caption{}\label{fig:means}
	\end{subfigure}
	~~
	\begin{subfigure}{0.2\textwidth}
		\includegraphics[height=1\textwidth]{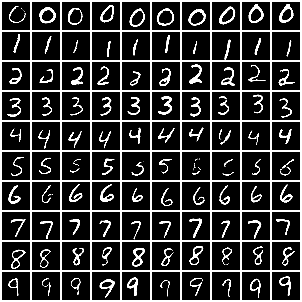}
		\caption{
		}\label{fig:samples}
	\end{subfigure}
	\caption{
		Visualizations of the latent space representations learned by supervised \method
    	on MNIST. 
        \textbf{(a)}: Images corresponding to means of the Gaussians corresponding to 
        different classes. 
        \textbf{(b)}: Class-conditional samples from the model at a reduced temperature $T=0.25$.}
        \label{fig:mean_samples}
\end{figure}

In Figure \ref{fig:means} we show the images $f^{-1}(\mu_i)$ corresponding
to the means of the Gaussians representing each class.
We see that the flow correctly learns to map the means to samples from the corresponding classes.
Next, in Figure \ref{fig:samples} we show class-conditional samples from the model. To produce a sample from class $i$, we first generate $z \sim \mN(\mu_i, T I)$,
where $T$ is a temperature parameter 
that controls trade-off between sample quality
and diversity; we then compute the samples as $f^{-1}(z)$. 
We set $T = 0.25^2$ to produce samples in Figure \ref{fig:samples}. 
As we can see, \method can produce reasonable class-conditional samples simultaneously
with achieving a high classification accuracy ($99.63\%$) on the MNIST dataset.

\end{document}